%% file: top.tex
\PassOptionsToPackage{table}{xcolor}
\documentclass[runningheads,table]{llncs}
\usepackage{graphicx}

\usepackage{tikz}
\usepackage{comment}
\usepackage{amsmath,amssymb} 
\usepackage{color}

\usepackage{booktabs}
\usepackage{tabularx}
\usepackage{overpic}
\usepackage{multirow}

\usepackage{caption}
\usepackage{subcaption}
\usepackage[pagebackref,breaklinks,colorlinks]{hyperref}

\usepackage[accsupp]{axessibility}  

\input{defs}

\begin{document}
\pagestyle{headings}
\mainmatter
\def\ECCVSubNumber{7511}  

\title{MeshUDF: Fast and Differentiable Meshing of Unsigned Distance Field Networks} 

\titlerunning{MeshUDF: Fast and Differentiable Meshing of UDF networks}
%
\author{Benoît Guillard\and
Federico Stella \and
Pascal Fua
}
\authorrunning{B. Guillard et al.}
%
\institute{CVLab, EPFL, \\
\email{benoit.guillard@epfl.ch} \email{federico.stella@epfl.ch} \email{pascal.fua@epfl.ch} }
\maketitle

\input{tex/0_abstract}

\input{tex/1_introduction}
\input{tex/2_related_work}
\input{tex/3_method}
\input{tex/4_experiments}
\input{tex/5_conclusion}

\subsubsection*{Acknowledgement.} This project was supported in part by the Swiss National Science Foundation.

\clearpage 
%
%
\bibliographystyle{splncs04}
\bibliography{string,vision,learning,cfd,graphics,optim,biomed,misc,local}

\clearpage
\input{tex/6_supp}

\end{document}

%% file: defs.tex
\newif\ifdraft
\draftfalse
\drafttrue

\usepackage[table]{xcolor}
\definecolor{orange}{rgb}{1,0.5,0}
\definecolor{violet}{RGB}{70,0,170}

\iffalse
 \newcommand{\PF}[1]{{\color{red}{\bf PF: #1}}}
 
 \newcommand{\BG}[1]{{\color{blue}{\bf BG: #1}}}
 
 \newcommand{\FS}[1]{{\color{orange}{\bf FS: #1}}}
 
\else
 \newcommand{\PF}[1]{}
 
 \newcommand{\BG}[1]{}
 
 \newcommand{\FS}[1]{}
 
\fi

\newcommand{\parag}[1]{\paragraph{#1}}

\newcommand{\bc}{\mathbf{c}}
\newcommand{\bx}{\mathbf{x}}
\newcommand{\bz}{\mathbf{z}}
\newcommand{\bv}{\mathbf{v}}
\newcommand{\bn}{\mathbf{n}}
\newcommand{\bgrad}{\mathbf{g}}

\newcommand{\bo}{\mathbf{o}}

\newcommand{\NDF}{\textit{NDF}}
\newcommand{\INFL}{\textit{Inflation}}

%% file: tex/0_abstract.tex

\begin{abstract}

Unsigned Distance Fields (UDFs) can be used to represent non-watertight surfaces. However, current approaches to converting them into explicit meshes tend to either be expensive or to degrade the accuracy. Here, we extend the marching cube algorithm to handle UDFs, both fast and accurately. Moreover, our approach to surface extraction is differentiable, which is key to using pretrained UDF networks to fit sparse data.

\end{abstract}

%% file: tex/1_introduction.tex

\input{figs/teaser.tex}

\section{Introduction}
\label{sec:introduction}

In recent years, deep implicit surfaces~\cite{Park19c,Mescheder19,Chen19c} have emerged as a powerful tool to represent and manipulate watertight surfaces. Furthermore, for applications that require an explicit 3D mesh, such as sophisticated rendering including complex physical properties~\cite{Nimier19} or optimizing physical performance~\cite{Baque18}, they can be used to parameterize explicit 3D meshes whose topology can change while preserving differentiability~\cite{Atzmon19,Remelli20b,Guillard22a}. However, these approaches can only handle watertight surfaces. Because common 3D datasets such as ShapeNet~\cite{Chang15} contain non-watertight meshes, one needs to preprocess them to create a watertight outer shell~\cite{Park19c,Xu19b}. This is time consuming and ignores potentially useful inner components, such as seats in a car. An alternative is to rely on network initialization or regularization techniques to directly learn from raw data~\cite{Atzmon20,Atzmon20b} but this significantly slows down the training procedure.

This therefore leaves open the problem of modeling non-watertight surfaces implictly. It has been shown in~\cite{Chibane20b,Zhao21a,Venkatesh21,Corona21} that occupancy fields and signed distance functions (SDFs) could be replaced by unsigned ones (UDFs) for this purpose.  However, unlike for SDFs, there are no fast algorithms to directly mesh UDFs. Hence, these methods rely on a two-step process that first extracts a dense point cloud that can then be triangulated using slow standard techniques~\cite{Bernardini99}. Alternatively, non-watertight surfaces can be represented as watertight thin ones surrounding them~\cite{Corona21,Guillard22a,Venkatesh21}. This amounts to meshing the $\epsilon$ iso-surface of an UDF using marching cubes~\cite{Lorensen87}, for $\epsilon$ being a small strictly positive scalar. Unfortunately, that degrades reconstruction accuracy because the thin surfaces cannot be infinitely so, as $\epsilon$ cannot be arbitrarily small. Furthermore, some applications such as draping simulation~\cite{Lahner18,Tang18d,Gundogdu22} require surfaces to be single-faced and cannot be used in conjunction with this approach.

In this paper, we first show that marching cubes can be extended to UDFs by reasoning on their gradients. When neighboring gradients face in opposite directions, this is evidence that a surface element should be inserted between them. We rely on this to replace the sign flips on which the traditional marching cube algorithm depends and introduce a new approach that exploits the gradients instead. This yields vertices and facets. When the UDF is parameterized by latent vectors, we then show that the 3D position of these vertices can be differentiated with respect to the latent vectors. This enables us to fit the output of pre-trained networks to sparse observations, such as 3D points on the surface of a target object or silhouettes of that object.

In short, our contribution is a new approach to meshing UDFs and parameterizing 3D meshes to model non-watertight surfaces whose topology can change while preserving differentiability, which is something that had only been achieved for watertight surfaces before. We use it in conjunction with a learned shape prior to optimize fitting to partial observations via gradient descent. We demonstrate it achieves better reconstruction accuracy than current deep-learning based approaches to handling non-watertight surfaces, in a fraction of the computation time, as illustrated by  Fig.~\ref{fig:teaser}. Our code is publicly available at \url{https://github.com/cvlab-epfl/MeshUDF/} .

%% file: figs/teaser.tex

\begin{figure}
	\begin{center}
	\begin{overpic}[width=.45\textwidth]{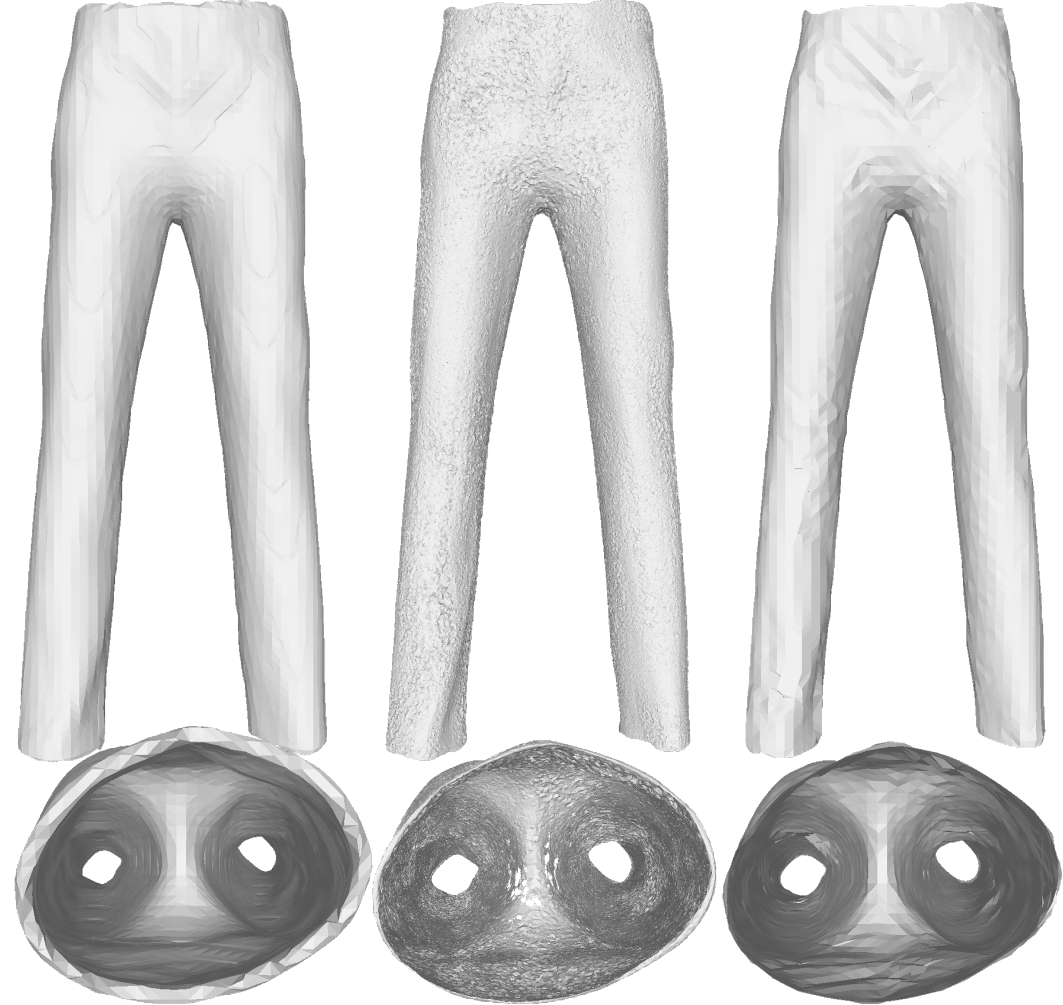}

		\put(15,-5){\small{(a)}}
		\put(48,-5){\small{(b)}}
		\put(81,-5){\small{(c)}}

	\end{overpic}
	\begin{overpic}[width=.47\textwidth, clip, trim={0cm -2.5cm 0cm 0}]{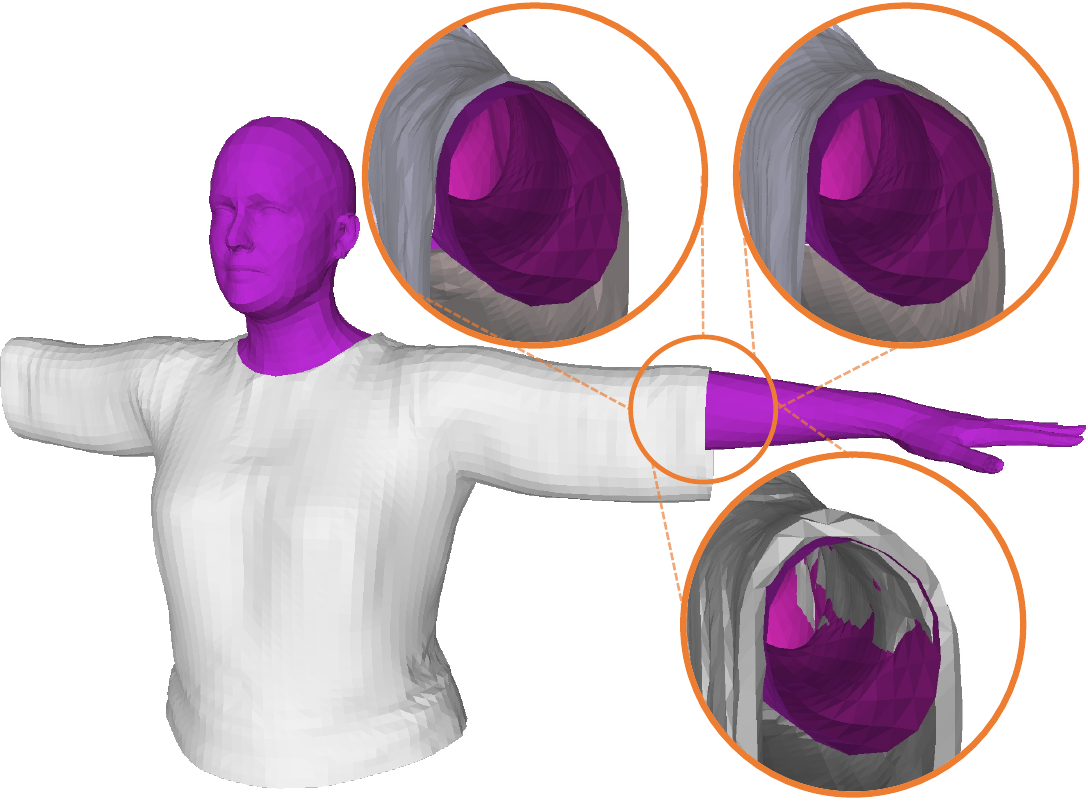}

		\put(48,-5){\small{(d)}}

	\end{overpic}
	\end{center}

		 \caption{\small \textbf{Meshing the UDF of a garment.} We present front and top views. \textbf{(a)} Inflating shapes to turn open surfaces into watertight ones~\cite{Venkatesh21,Corona21,Guillard22a} inherently reduces accuracy by making the surface thicker, as shown in top view. \textbf{(b)} Triangulating a cloud of 3D points collapsed on the $0$-levelset~\cite{Chibane20b} is time-consuming and tends to produce rough surfaces. \textbf{(c)} Directly meshing the UDF using our approach is more accurate and less likely to produce artifacts. In addition, it makes the iso-surface extraction process differentiable. \textbf{(d}) We mesh the UDF of a shirt and display it on a human body.
		 The three insets represent the ground truth shirt, the reconstruction with our method, and the inflation approach, respectively. Our approach---in the upper right inset---produces fewer artifacts and no penetrations with the body.}
	\label{fig:teaser}
\end{figure}

%% file: tex/2_related_work.tex

\section{Related Work}
\label{sec:related}

Deep implicit surfaces~\cite{Park19c,Mescheder19,Chen19c} have proved extremely useful to model watertight surfaces using occupancy grids and SDFs, while non-watertight surfaces can be handled either using UDFs or inflating an SDF around them. As meshing almost always involves using a version of the classic marching cube algorithm~\cite{Lorensen87}, we discuss these first. We then review recent approaches to representing non-watertight surfaces using implicit surfaces.

\parag{Triangulating an Implicit Field.}

Marching cubes was originally proposed in~\cite{Lorensen87} and refined in~\cite{Chernyaev95,Lopes03,Lewiner03} to triangulate the $0$-isosurface of a 3D scalar field. It marches sequentially across cubic grid cells and if field values at neighboring corners have opposing signs, triangular facets are created according to a manually defined lookup table. Vertices of these triangle facets are adjusted by linear interpolation over the field values. Since then, newer methods have been developed such as dual methods~\cite{Ju02}. They are better at triangulating surfaces with sharp edges at the expense of increased complexity and requiring a richer input. Hence, due to its simplicity and flexibility, along with the availability of efficient implementations, the original algorithm of~\cite{Lewiner03} remains in wide use~\cite{Mescheder19,Park19c,Peng20c,Hao20a,Xu19b}. More recently,~\cite{Chen21c} proposed a data driven approach at improving sharp features reconstructed by marching cubes. Even though marching cubes is not differentiable, it has been shown that gradients can be estimated for surface points, thus allowing backpropagation~\cite{Atzmon19,Remelli20b}. However, this approach requires surface normals whose orientation is unambiguous, which makes it impractical when dealing with non-watertight surfaces.

\parag{Triangulating Implicit Non-Watertight Surfaces.}

Unfortunately, neither the original marching cubes algorithm nor any of its recent improvements are designed to handle non-watertight surfaces. One way around this is to surround the target surface with a thin watertight one~\cite{Corona21,Guillard22a,Venkatesh21}, as shown in Fig.~\ref{fig:teaser}(a). One can think of the process as {\it inflating} a watertight surface around the original one. Marching cubes can then be used to triangulate the inflated surface, but the result will be some distance away from the target surface, resulting in a loss of accuracy. Another approach is to sample the minimum level set of an UDF field, as in NDF~\cite{Chibane20b} and AnchorUDF~\cite{Zhao21a}. This is done by projecting randomly initialized points on the surface using gradient descent. To ensure full coverage, points are re-sampled and perturbed during the process. This produces a cloud, but not a triangulated mesh with information about the connectivity of neighboring points. Then the ball-pivoting method~\cite{Bernardini99}, which connects neighboring points one triplet at a time, is used to mesh the cloud, as shown in Fig.~\ref{fig:teaser}(b).  It is slow and inherently sequential.

%% file: tex/3_method.tex

\section{Method}
\label{sec:method}

We now present our core contribution, a fast and differentiable approach to extracting triangulated isosurfaces from unsigned distance fields produced by a neural network. Let us consider  a network that implements a function
\begin{align}
  \phi: \mathbb{R}^C \times \mathbb{R}^3 &\rightarrow \mathbb{R}^+ \; , \\
  \bz, \bx &\mapsto s \nonumber \; ,
\end{align}
where $\bz \in  \mathbb{R}^C$ is a parameter vector; $\bx$ is a 3D point; $s$ is the Euclidean distance to a surface. Depending on the application, $\bz$ can either represent only a latent code that parameterizes the surface or be the concatenation of such a code and the network parameters. In Sec.~\ref{sec:mesh}, we propose an approach to creating a triangulated mesh $M=(V,F)$ with vertices $V$ and facets $F$ from the $0$-levelset of the scalar field $\phi(\bz, \cdot)$. Note that it could also apply to non-learned UDFs, as shown in the supplementary material. In Sec.~\ref{sec:differentiating}, we show how to make the vertex coordinates differentiable with respect to $\bz$. This allows refinement of shape codes or network parameters with losses directly defined on the mesh.

\subsection{From UDF to Triangulated Mesh}
\label{sec:mesh}

\subsubsection{Surface Detection within Cells}
\label{sec:detection}

As in standard marching cubes~\cite{Lorensen87}, we first sample a discrete regular grid $G$ in the region of interest, typically $[-1,1]^3$. At each location $\bx_i \in G$ we compute
\begin{align}
  u_i  = \phi(\bz, \bx_i) \; , \; \; \; \; \; \; \; \; \; \;
  \bgrad_i = \nabla_{\bx}\phi(\bz, \bx_i)  \; , \nonumber
\end{align}
where $u_i$ is the unsigned distance to the implicit surface at location $\bx_i$, and $\bgrad_i \in \mathbb{R}^3$ is the gradient computed using backpropagation. Given a cubic cell and its 8 corners, let $(u_1,...,u_8)$, $(\bx_1,...,\bx_8)$, and $(\bgrad_1,...,\bgrad_8)$ be the above values in each one. Since all $u_i$ are positive, a surface traversing a cell does not produce a sign flip as it does when using an SDF. However, when corners $\bx_i$ and $\bx_j$ lie on opposite sides of the $0$-levelset surface, their corresponding vectors $\bgrad_i$ and $\bgrad_j$ should have opposite orientations, provided the surface is sufficiently smooth within the cell. Hence, we define a \textit{pseudo-signed distance}
\begin{align}
s_i = \text{sgn}(\bgrad_1 \cdot \bgrad_i) u_i \; ,
\label{eq:pseudo_signed_dist}
\end{align}
where $\bx_1$ is one of the cell corners that we refer to as the {\it anchor}. $\bx_1$ is assigned a positive pseudo-signed distance and corners where the gradient direction is opposite to that at $\bx_1$ a negative one. When there is at least one negative $s_i$, we use marching cubes' disjunction cases and vertex interpolation to reconstruct a triangulated surface in the cell. Computing pseudo-signs in this way is simple but has two shortcomings.  First, it treats each cell independently, which may cause inconsistencies in the reconstructed facets orientations. Second, especially when using learned UDF fields that can be noisy~\cite{Venkatesh21}, the above smoothness assumption may not hold within the cells. This typically results in holes in the reconstructed meshes.
\input{figs/method_AND_artefact2.tex}

To mitigate the first problem, our algorithm starts by exploring the 3D grid until it finds a cell with at least one negative pseudo-sign. It then uses it as the starting point for a breadth-first exploration of the surface. Values computed at any cell corner are stored and never recomputed, which ensures that the normal directions and interpolated vertices are consistent in adjacent cells. The process is repeated to find other non-connected surfaces, if any. To mitigate the second problem we developed a more sophisticated method to assign a sign to each cell corner. We do so as described above for the root cell of our breadth-first search, but we use the voting scheme depicted by Fig.~\ref{fig:method} for the subsequent ones. Voting is used to aggregate information from neighboring nodes to estimate pseudo-sign more robustly.
Each corner $\bx_i$ of a cell under consideration receives votes from all adjacent grid points $\bx_k$ that have already been assigned a pseudo-sign, based on the relative directions of their gradients and the pseudo-sign of $\bx_k$. Since gradients locally point towards the greatest ascent direction, if the projections of $\bgrad_i$ and $\bgrad_k$ along the edge connecting $\bx_i$ and $\bx_k$ face each other, there is no surface between them and the vote is in favor of them having the same sign: $v_{ik} = \text{sgn}(s_k)$. Otherwise, the vote depends on gradient directions and we take it to be
\begin{align}
  \label{eq:voting_scheme}
  v_{ik} = (\bgrad_i \cdot \bgrad_k) \text{sgn}(s_k)
\end{align}
because the more the gradients are aligned, the more confident we are about the two points being on the same side of the surface or not, depending on the sign of the dot product. The sign of the sum of the votes is assigned to the corner.

If one of the $\bx_k$ is zero-valued its vote does not contribute to the scheme, but it means that there is a clear surface crossing. This can happen when meshing learned UDF fields at higher resolutions, because their 0-level set can have a non-negligible volume. Thus, the first non-zero grid point along its direction takes its place in the voting scheme, provided that it has already been explored. To further increase the reliability of these estimates, grid points with many disagreeing votes are put into a lower priority queue to be considered later, when more nearby grid points have been evaluated and can help produce a more consistent sign estimate. In practice we only perform these computations within cells whose average UDF values of $(u_1,...,u_8)$ are small. Others can be ignored, thus saving computation time and filtering bad cell candidates which have opposing gradients but lie far from the surface.

\parag{Global Surface Triangulation}
\label{sec:artefacts}

The facets that the above approach yields are experimentally consistent almost everywhere, except for a small number of them, which we describe below and can easily remove in a post-processing stage. Note that the gradients we derive in Sec.~\ref{sec:differentiating} do not require backpropagation through the iso-surface extraction. Hence, this post-processing step does not compromise differentiability.

\parag{Removing Spurious Facets and Smoothing Borders.}

As shown in  Fig.~\ref{fig:artefact}~\textbf{(a)}, facets that do not correspond to any part of the surfaces can be created in cells with gradients pointing in opposite directions without intersecting the $0$-levelset. This typically happens near surface borders because our approach tends to slightly extend them, or around areas with poorly approximated gradients far from the surface in the case of learned UDF fields. Such facets can be detected by re-evaluating the distance field on all vertices. If the distance field for one vertex of a face is greater than half the side-length of a cubic cell, it is then eliminated. Moreover, since marching cubes was designed to reconstruct watertight surfaces, it cannot handle surface borders. As a result, they appear slightly jagged on initial reconstructions. To mitigate this, we apply Laplacian smoothing on the edges belonging to a single triangle. This smoothes borders and qualitatively improves reconstructions, as shown in Fig.~\ref{fig:artefact}~\textbf{(b)}.


\subsection{Differentiating through Iso-Surface Extraction}
\label{sec:differentiating}

Let $\bv \in \mathbb{R}^3$ be a point on a facet reconstructed using the method of Sec.~\ref{sec:mesh}. Even though differentiating $\bv$ directly through marching cubes is not possible~\cite{Liao18a,Remelli20b}, it was shown that if $\phi$ were an SDF instead of an UDF, derivatives could be obtained by reasoning about surface inflation and deflation~\cite{Atzmon19,Remelli20b}.
Unfortunately, for an UDF, there is no "in" or "out" and its derivative is undefined on the surface itself. Hence, this method does not directly apply. Here we extend it so that it does, first for points strictly within the surface, and then for points along its boundary.

\input{figs/method_diff_AND_border.tex}
\textit{Derivatives within the Surface.}
Let us assume that $\bv \in \mathbb{R}^3$ lies within a facet where the surface normal $\bn$ is unambiguously defined up to its orientation. Let us pick a small scalar value $\alpha > 0$ and consider
\begin{align}
  \bv_+ = \bv + \alpha \bn  \; \text{ and } \;  \bv_- = \bv - \alpha \bn \; , \nonumber
\end{align}
the two closest points to $\bv$ on the $\alpha$-levelset on both sides of the 0-levelset. For $\alpha$ small enough, the outward oriented normals at these two points are close to being $\bn$ and $-\bn$. We can therefore use the formulation of~\cite{Atzmon19,Remelli20b} to write
\begin{align}
  \frac{\partial \bv_+}{\partial \bz} \approx - \bn \frac{\partial \phi}{\partial \bz}(\bz, \bv_+) \; \; \; \; \; \; \text{ and } \; \; \; \; \; \; \; \frac{\partial \bv_-}{\partial \bz} \approx  \bn \frac{\partial \phi}{\partial \bz}(\bz, \bv_-)\; .
  \label{eq:derivatives_vPlusMinus}
\end{align}
Since $\bv = \tfrac{1}{2} (\bv_- + \bv_+)$, Eq.~\ref{eq:derivatives_vPlusMinus}, yields
\begin{align}
\frac{\partial \bv}{\partial \bz} &\approx  \frac{\bn}{2} \left [ \frac{\partial \phi}{\partial \bz}(\bz, \bv - \alpha \bn) - \frac{\partial \phi}{\partial \bz}(\bz, \bv + \alpha \bn)\right ]   \; . \label{eq:derivatives0}
\end{align}
We provide a more formal proof  and discuss the validity of the approximation in appendix. Note that using $\bn'=-\bn$ instead of $\bn$ yields the same result. Intuitively, this amounts to surrounding the 0-levelset with $\alpha$-margins where UDF values can be increased on one side and decreased on the other, which allows local deformations perpendicular to the surface. Fig.~\ref{fig:method_diff}~\textbf{(a)} depicts the arrangement of $\bv$, $\bv_+$ and $\bv_-$ around the surface. The derivative of Eq.~\ref{eq:derivatives0} implies that infinitesimally increasing the UDF value at $\bv_-$ and decreasing it at $\bv_+$ would push $\bv$ in the direction of $\bn$, as shown in Fig.~\ref{fig:method_diff} \textbf{(b)}, and conversely. In practice, we use $\alpha=10^{-2}$ in all our experiments.

\textit{Derivatives at the Surface Boundaries.} Let us now assume that $\bv$ sits on the edge of a boundary facet. Mapping it to $\bv_+$ and $\bv_-$ and using the derivatives of Eq.~\ref{eq:derivatives0} would mean that all deformations are perpendicular to that facet. Thus, it does not permit shrinking or expanding of the surface during shape optimization. In this setting, there is a whole family of closest points to $\bv$ in the $\alpha$-levelset;  they lay on a semicircle with radius $\alpha$. To allow for shrinkage and expansion, we map $\bv$ to the semicircle point along $\bo$, a vector perpendicular to the edge, pointing outwards, and within the plane defined by the facet. Hence, we consider the point $\bv_o$ which is the closest to $\bv$ on the $\alpha$-levelset in the direction of $\bo$:
\begin{align}
\bv_o &= \bv + \alpha \bo
\label{eq:mapping_Out}
\end{align}
For $\alpha$ small enough, the outward oriented normal at $\bv_o$ is $\bo$ and we again use the formulation of~\cite{Atzmon19,Remelli20b} and Eq.~\ref{eq:mapping_Out} to write
\begin{align}
  \frac{\partial \bv_o}{\partial \bz} = - \bo \frac{\partial \phi}{\partial \bz}(\bz, \bv_o) \; \text{ and } \; \frac{\partial \bv}{\partial \bz}  = - \bo \frac{\partial \phi}{\partial \bz}(\bz, \bv + \alpha \bo) \; , \label{eq:derivatives5}
\end{align}
which we use for all points $\bv$ on border edges. As shown in Fig.~\ref{fig:method_diff_border}, this implies that increasing the UDF value at $\bv_o$ would push $\bv$ inwards and make the surface shrink. Conversely, decreasing it extends the surface in the direction of $\bo$.

%% file: figs/method_AND_artefact2.tex

\begin{figure}[t]
	\centering
	\begin{minipage}{.52\textwidth}
		\centering
		\begin{overpic}[width=\textwidth]{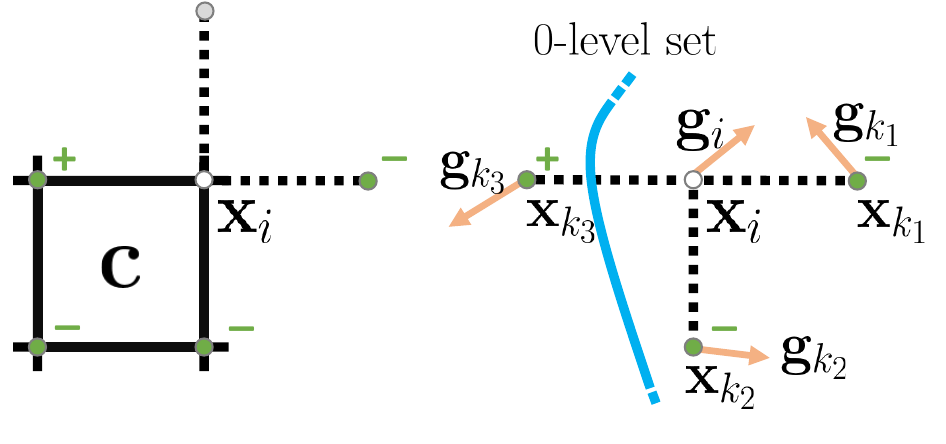}
			\put(25,-0.5){\small{(a)}}
			\put(70,-0.5){\small{(b)}}
		\end{overpic}
		\captionof{figure}{\small \textbf{Voting.} \textbf{(a)} Corner $\bx_i$ of cell $\bc$ has 3 neighbors that already have a pseudo-sign  and vote. \textbf{(b)} The projections of $\bgrad_i$ and $\bgrad_{k_1}$ on the edge connecting the two neighbors face each other. Thus $\bx_{k_1}$ votes for $\bx_i$ having the same sign as itself (-). The other two neighbors vote for $-$ as well given the result of computing Eq.~\ref{eq:voting_scheme}.}
		\label{fig:method}
	\end{minipage}%
	$ $
	\begin{minipage}{.45\textwidth} 
		\centering
		\begin{overpic}[width=\textwidth]{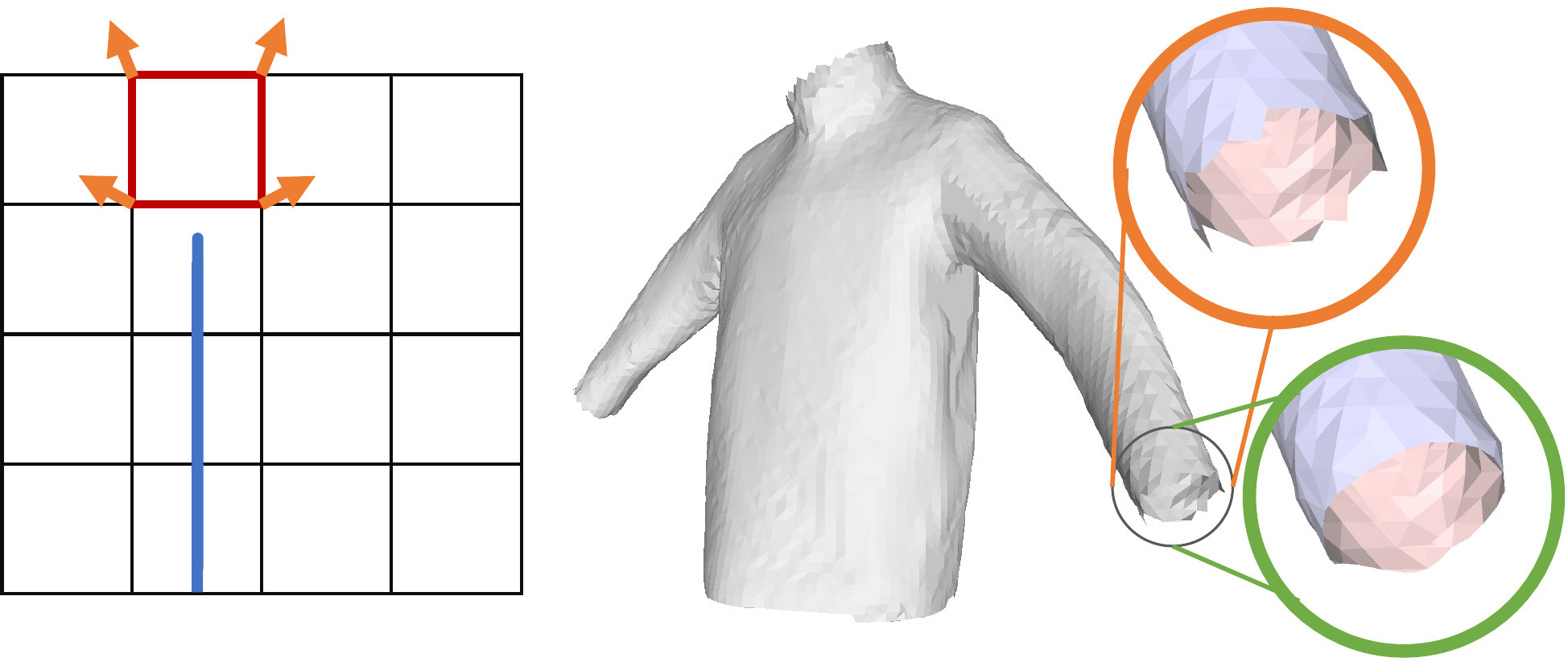}
			\put(15,0){\small{(a)}}
			\put(68,0){\small{(b)}}
		\end{overpic}
		\captionof{figure}{\small \textbf{Removing artifacts.} \textbf{(a)} Given the blue $0$-level surface, the red cell has gradients in opposing directions and yields an undesirable face. We prune these by evaluating the UDF on reconstructed faces. \textbf{(b)} Initially reconstructed borders are uneven (top). We smooth them during post-processing (bottom).}
		\label{fig:artefact}
	\end{minipage}
\end{figure}

%% file: figs/method_diff_AND_border.tex

\begin{figure}[t]
	\centering
	\begin{minipage}{.485\textwidth}
		\centering
		\begin{overpic}[width=\textwidth]{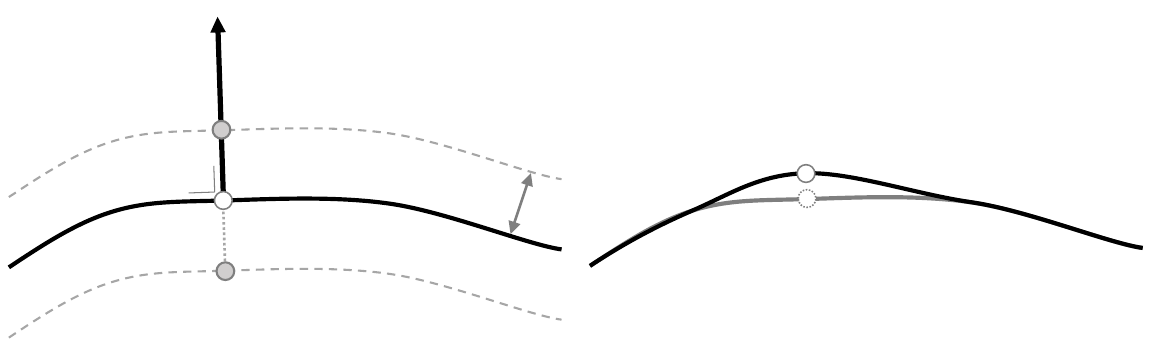}
			\put(15,27){\small{$\bn$}}
			\put(20.3,8){\small{$\bv_-$}}
			\put(20,14.5){\small{$\bv$}}
			\put(19.8,20.4){\small{$\bv_+$}}
			\put(45.5,11.5){\footnotesize{$\alpha$}}

			\put(0.5,6){\textcolor{darkgray}{\scriptsize$\text{=}0$}}
			\put(1,14){\textcolor{darkgray}{\scriptsize$>\!\!0$}}
			\put(0.5,0){\textcolor{darkgray}{\scriptsize$>\!\!0$}}

			\put(22.5,-2){\small{(a)}}
			\put(72.5,-2){\small{(b)}}

			\put(71,17){\small{$\bv'$}}
			\put(71,10.5){\textcolor{gray}{\small{$\bv$}}}

		\end{overpic}
		\captionof{figure}{\textbf{Iso-surface deformation}: \textbf{(a)} $\bv$ on the $0$-levelset, $\bv_+$ and $\bv_-$ at distance $\alpha$ ; \textbf{(b)} $\bv$ moves to $\bv'$ if the UDF decreases at $\bv_+$ and increases at $\bv_-$.}
		\label{fig:method_diff}
	\end{minipage}%
	$ $
	\begin{minipage}{.485\textwidth} 
		\centering
		\begin{overpic}[clip, trim={0cm 0.3cm 0 0.cm},width=\textwidth]{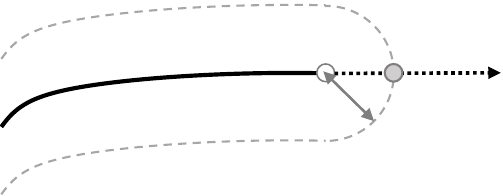}

			\put(28,10.5){\small{$\bv$}}
			\put(34,10.5){\small{$\bv_o$}}
			\put(41,5){\small{$\bo$}}
			\put(32.5,3){\footnotesize{$\alpha$}}
			\put(75,10.5){\small{$\bv'$}}
			\put(87,10.5){\textcolor{gray}{\small{$\bv$}}}

			\put(3.5,3.5){\textcolor{darkgray}{\tiny$\text{=}0$}}
			\put(3,10.5){\textcolor{darkgray}{\tiny$>\!\!0$}}

			\put(22.5,-5.5){\small{(a)}}
			\put(72.5,-5.5){\small{(b)}}

		\end{overpic}
		\captionof{figure}{\textbf{Iso-surface shrinkage or extension}: \textbf{(a)} with $\bv$ on the border of the $0$-levelset, we place $\bv_o$ at a distance $\alpha$ in the direction of $\bo$ ; \textbf{(b)} If the UDF increases at $\bv_o$, $\bv$ moves to $\bv'$.}
		\label{fig:method_diff_border}
	\end{minipage}
\end{figure}

%% file: tex/4_experiments.tex

\section{Experiments}
\label{sec:experiments}

We demonstrate our ability to mesh UDFs created by deep neural networks. To this end, we first train a deep network to map latent vectors to UDFs representing different garments, that is, complex open surfaces with many different topologies. We then show that, given this network, our approach can be used to effectively triangulate these garments and to model previously unseen ones. Next, we plug our triangulation scheme into existing UDF networks and show that it is a straightforward operation. Finally, the benefit of the border gradients of Sec.~\ref{sec:differentiating} is evaluated. The voting scheme proposed in Sec.~\ref{sec:mesh} is ablated in appendix, where more qualitative results are also shown.

\subsection{Network and Metrics}

Our approach is designed to triangulate the output of networks that have been trained to produce UDF fields. To demonstrate this, we use an auto-encoding approach~\cite{Park19c} with direct supervision on UDF samples on the MGN dataset~\cite{Bhatnagar19}  to train a network $\phi_{\theta}$  that maps latent vectors of dimension 128 to UDFs that represent garments. These UDFs can in turn be triangulated using our algorithm to produce meshes such as those of Fig.~\ref{fig:teaser}. We provide details of this training procedure in the supplementary material. The MGN dataset comprises 328 meshes. We use 300 to train $\phi_{\theta}$ and the remaining 28 for testing. For comparison purposes, we also use the publicly available pre-trained network of NDF~\cite{Chibane20b} that regresses UDF from sparse input point clouds. It was trained on raw ShapeNet~\cite{Chang15} meshes, without pre-processing to remove inner components make them watertight or consistently orient facets.

To compare the meshes we obtain to the ground-truth ones, we evaluate the following three metrics (details in the supplementary material):
%
\begin{itemize}

  \item The \textbf{Chamfer distance} (CHD) measures the proximity of 3D points sampled from the surfaces, the lower the better.

  \item The \textbf{Image consistency} (IC) is the product of IoU and cosine similarity of 2D renderings of normal maps from 8 viewpoints, the higher the better.

  \item The \textbf{Normal consistency} (NC) quantifies the agreement of surface normals in 3D space, the higher the better.

\end{itemize}
%

\subsection{Mesh Quality and Triangulation Speed}
\label{subsec:accuracy_speed}

Fig.~\ref{fig:teaser} was created by triangulating a UDF produced by $\phi_{\theta}$ using either our meshing procedure (\textit{Ours}) or one of two baselines:
%
\begin{itemize}

 \item \textit{BP}. It applies the ball-pivoting method~\cite{Bernardini99} implemented in~\cite{Cignoni08} on a dense surface sampling of $900k$ points, as originally proposed in~\cite{Chibane20b} and also used in~\cite{Zhao21a}. Surface points are obtained by gradient descent on the UDF field.

 \item \textit{Inflation}~\cite{Guillard22a,Venkatesh21,Corona21}.  It uses standard marching cubes to mesh the $\epsilon$-isolevel of the field, with $\epsilon > 0$.

\end{itemize}
%
In Tab.~\ref{tab:meshing_comparison} (left), we report metrics on the 300 UDF fields $\phi_{\theta}(\mathbf{z}_i, \cdot)$ for which we have latent codes resulting from the above-mentioned training.
\textit{Inflation} and \textit{Ours} both use a grid size of $128^3$ over the $[-1, 1]^3$ bounding box, and we set \textit{Inflation}'s $\epsilon$ to be 55\% of marching cubes' step size.
In Tab.~\ref{tab:meshing_comparison} (right) we also report metrics for the pretrained NDF network~\cite{Chibane20b} tested on 300 ShapeNet cars, in which case we increase \textit{Inflation} and \textit{Ours} resolution to $192^3$ to account for more detailed shapes. An example is shown in Fig.~\ref{fig:ndf_anchorudf} The experiments were run on a NVidia V100 GPU with an Intel Xeon 6240 CPU.

\input{tables/meshing_comparison.tex}

As shown on the left of Table.~\ref{tab:meshing_comparison},  \textit{Ours} is slightly more accurate than \NDF~in terms of all three metrics, while being orders of magnitude faster.  \textit{Inflation} is even faster---this reflects the overhead our modified marching cube algorithm imposes---but far less accurate. To show that this result is not specific to garments, we repeated the same experiment on 300 cars from the ShapeNet dataset and report the results on the right side of Table.~\ref{tab:meshing_comparison}. The pattern is the same except for NC, which is slightly better for \textit{Inflation}. We conjecture this to be a byproduct of the smoothing provided by \textit{Inflation}, which is clearly visible in Fig.~\ref{fig:teaser}\textbf{(a,c)}. To demonstrate that these results do not depend on the specific marching cube grid resolution we chose, we repeated the experiment for grid resolutions ranging from 64 to 512 and plot the average CHD as a function of resolution in Fig.~\ref{fig:CHD_MC_res}. It remains stable over the whole range. For comparison purposes, we also repeated the experiment with \textit{Inflation}. Each time we increase the resolution, we take the $\epsilon$ value that defines the iso-surface to be triangulated to be 10\% greater than half the grid-size, as shown in Fig.~\ref{fig:choice_eps}. At very high resolution, the accuracy of \INFL{} approaches \textit{Ours} but that also comes at a high-computational cost because operating on $512 \times 512 \times 512$ cubes instead of  $128 \times 128 \times 128$ ones is much slower, even when using multi-resolution techniques.

\input{figs/choice_eps_AND_chd_mc_res.tex}

\subsection{Using Differentiability to Fit Sparse Data}
\label{subsec:differentiability_demos}

Given the trained network $\phi_{\theta}$ and latent codes for training shapes from Sec.~\ref{subsec:accuracy_speed}, we now turn to recovering latent codes for the remaining 28 test garments. For each test garment $G_j$, given the UDF representing it, this would be a simple matter of minimizing the mean square error between it and the field $\phi_{\theta}(\bz,\cdot)$ with respect to $\bz$, which does not require triangulating. We therefore consider the more challenging and more realistic cases where we are only given eiter small set of 3D points $P_j$---in practice we use 200 points---or silhouettes and have to find a latent vector that generates the UDF that best approximates them.

\parag{Fitting to 3D points.}
\label{subsubsec:differentiability_demos_PC}
\input{tables/optim_PC.tex}
One way to do this is to remain in the implicit domain and to minimize one of the two loss functions
\begin{align}
\mathcal{L}_{PC,UDF}(P_j, \mathbf{z}) &= \tfrac{1}{|P_j|} \sum_{p \in {P_j}} | \phi_{\theta}(\mathbf{z}, p) | \; , \\
 \widetilde{\mathcal{L}}_{PC,UDF}(P_j, \mathbf{z}) & = \mathcal{L}_{PC,UDF}(P_j, \mathbf{z})
	 + \tfrac{1}{|A|} \sum_{a \in A} | \phi_{\theta}(\mathbf{z}, a) - \min_{p \in {P_j}} \left \| a - p \right \|_2 | \; , \nonumber
\end{align}
where $A$ is a set of randomly sampled points. Minimizing $\mathcal{L}_{PC,UDF}$ means that the given $P_j$ points must be on the zero-level surface of the UDF.
Minimizing $\widetilde{\mathcal{L}}_{PC,UDF}$ means that, in addition, the predicted UDF evaluated at points of $A$ must match the approximated UDF computed from $P_j$. Since the latter is sparse, $\widetilde{\mathcal{L}}_{PC,UDF}$ only provides an approximate supervision.

An alternative is to use our approach to triangulate the UDFs  and minimize the loss function
\begin{equation}
\mathcal{L}_{PC,mesh}(P_j, \bz) = \tfrac{1}{|P_j|} \sum_{p \in {P_j}} \min_{a \in M_{\mathbf{z}}} \left \| a - p \right \|_2 \; ,
\label{eq:chdLoss}
\end{equation}
where $a \in M_{\mathbf{z}}$ means sampling 10k points $a$ on the triangulate surface of $M_{\mathbf{z}}$. Minimizing $\mathcal{L}_{PC,mesh}$ means that the chamfer distance between the triangulated surfaces and the sample points should be small. Crucially, the results of Sec.~\ref{sec:differentiating} guarantee that  $\mathcal{L}_{PC,mesh}$ is differentiable with respect to $\bz$, which makes minimization practical. We tried minimizing the three loss functions defined above. In each case we started the minimization from a randomly chosen latent vector for a garment of the same type as the one we are trying to model, which corresponds to a realistic scenario if the initial estimate is provided by an upstream network. We report our results in Tab.~\ref{tab:optim_PC}. Minimizing $\mathcal{L}_{PC,mesh}$ clearly yields the best results, which highlights the usefulness of being able to triangulate and to differentiate the result.

\parag{Fitting to Silhouettes.}
\label{subsubsec:differentiability_demos_silh}

\input{tables/optim_silh.tex}

We now turn to the problem of fitting garments to rasterized binary silhouettes. Each test garment $j$ is rendered into a front-facing binary silhouette $S_j \in \{0,1\}^{256 \times 256}$. Given $S_j$ only, our goal is to find the latent code $\mathbf{z}_j$ that best encodes $j$. To this end, we minimize
\begin{equation}
\mathcal{L}_{silh,mesh}(S_j, \bz) = L_1(rend(M_{\mathbf{z}}), S_j) \; ,
\end{equation}
where $rend$ is a differentiable renderer~\cite{Kato18} that produces a binary image of the UDF triangulation $M_{\bz}$ and $L_1(\cdot)$ is the $L_1$ distance.  Once again, the differentiability of $M_{\bz}$ with respect to $\bz$ is key to making this minimization practical.

In theory, instead of rendering a triangulation, we could have used an UDF differential renderer. Unfortunately, we are not aware of any.  Approaches such as that of~\cite{Liu20b}  rely on finding sign changes and only work with SDFs. In contrast, CSP-Net~\cite{Venkatesh21} can render UDFs without meshing them but is not differentiable. To provide a baseline, we re-implemented SMPLicit's strategy~\cite{Corona21} for fitting a binary silhouette by directly supervising UDF values. We sample a set of points $P \subset [-1,1]^3$, and project each $p \in P$ to $S_j$ using the front-facing camera $\mathbf{c}$ to get its projected value $s_p$. If $s_p=1$, point $p$ falls within the target silhouette, otherwise it falls into the background. SMPLicit's authors advocate optimizing $\mathbf{z}$ by summing
\begin{equation}
\mathcal{L}_{silh,UDF}(S_j, \mathbf{z}) = \left\{\begin{matrix}
  |\phi_{\theta}(\mathbf{z}, p) - d_{max}| & \mbox{if }s_p=0\\
  \underset{\bar{p}\mbox{ s.t. } \mathbf{c}(\bar{p})=\mathbf{c}(p)}{\min} |\phi_{\theta}(\mathbf{z}, \bar{p})| & \mbox{if }s_p=1
  \end{matrix}\right.  \; .
 \end{equation}
on $p\in P$. That is, points projecting outside the silhouette ($s_p=0$) should have a UDF value equal to the clamping value $d_{max}$. For points projecting inside the silhouette, along a camera ray we only consider $\bar{p}$, the closest point to the current garment surface estimate and its predicted UDF value should be close to $0$. We report our results in Tab.~\ref{tab:optim_silh}. Minimizing $\mathcal{L}_{silh,mesh}$ yields the best results, which highlights the benefits of pairing our method with a differentiable mesh renderer.

\input{tables/ablate_grads.tex}
\textit{Ablation Study.} \label{sec:ablation}
We re-ran the optimizations without the border derivative term of Eq.~\ref{eq:derivatives5}, that is, by computing the derivatives everywhere using the expression of Eq.~\ref{eq:derivatives0}. As can be seen in  Tab.~\ref{tab:ablate_grads}, this reduces performance and confirms the importance of allowing for shrinkage and expansion of the garments.

\subsection{Differentiable Topology Change}
\label{subsec:change_topology}

\input{figs/topology_change.tex}

A key feature of all implicit surface representations is that they can represent surfaces whose topology can change. As shown in Fig.~\ref{fig:topology_change}, our approach allows us to take advantage of this while simultaneously creating a mesh whose vertices have associated spatial derivatives. To create this example, we started from a latent code for a pair of pants and optimized with respect to it to create a new surface that approximates a sweater by minimizing the CHD loss of~Eq.~\ref{eq:chdLoss} over
10k 3D points on that sweater. The topology changes that occur on the mesh representing the deforming shape do not create any difficulties.

\subsection{Generalization to other UDF Networks}
\label{subsec:generalization}

\input{figs/ndf_anchorudf.tex}

To show that our meshing procedure is applicable as-is to other UDF-based methods, we use it downstream of publicly available pre-trained networks. In Fig.~\ref{fig:ndf_anchorudf} (bottom) we mesh the outputs of the garment reconstruction network of AnchorUDF~\cite{Zhao21a}. In Fig.~\ref{fig:ndf_anchorudf} (top) we apply it to the point cloud completion pipeline of NDF~\cite{Chibane20b}. Both these methods output dense point clouds surface, which must then be meshed using the time-costly ball pivoting algorithm. Instead, our method can directly mesh the UDF and does so in a fraction of the time while preserving differentiability. That makes the whole algorithm suitable for inclusion into an end-to-end differentiable pipeline.

\subsection{Limitations}
\label{subsec:limitations}
\textit{Reliance on learned UDF fields.}
The proposed method can mesh the zero-surface of an unsigned distance field. In practice however, UDF fields are approximated with neural networks, and we find it difficult to learn a sharp $0$-valued surface for networks with small capacities. It can for example happen that the approximate UDF field is not reaching zero, or that the zero surface thickens and becomes a volume, or that the gradients are not approximated well enough. In such cases, artifacts such as single-cell holes can appear when using our method at a high resolution. Note that applying our method to a real UDF would not exhibit such issues. By comparison however, applying marching cubes on an approximate and poorly learned SDF is more robust since it only requires the field to be continuous and to have a zero crossing to produce artifact-free surfaces. UDF networks could be made more accurate by using additional loss terms~\cite{Gropp20} or an adaptive training procedure~\cite{Duan20}, but this research direction is orthogonal to the method proposed in this paper. Moreover, similarly to~\cite{Remelli20b} for SDFs, since the proposed gradients rely on the field being an UDF, they cannot be used to train a neural network from scratch.
This would require network initialization or regularization strategies to ensure it regresses valid UDF fields, a topic we see as an interesting research direction.


\textit{Limitations of marching cubes.}
After locally detecting surface crossings via the pseudo-sign computation, we rely on standard marching cubes for meshing an open surface, which implies the need of a high resolution grid to detect high frequency details, and cubic scalability over grid resolution. Moreover, marching cubes was designed to handle watertight surfaces, and as a consequence some topological cases are missing, for example at surface borders or intersections. This could be remedied by detecting and handling such new cases with additional disjunctions. Finally, the breadth-first exploration of the surface makes the orientation of adjacent facets consistent with each other. However, non-orientable surfaces such as M\"{o}bius-strips would intrisically produce juncture points with inconsistent orientations when two different branches of the exploration reach each other. In such points, our method can produce holes. Similarly, marching cubes has geometric guarantees on the topology of reconstructed meshes, but this is not true for the proposed method since there is no concept of \textit{inside} and \textit{outside} in UDFs.


%% file: tables/meshing_comparison.tex

\begin{table}[t]
	\caption{\small \textbf{Comparing UDF meshing methods.} Average Chamfer (CHD), image consistency (IC), normal consistency (NC) and processing time for 300 garments (left) and 300 ShapeNet cars (right). We use a single UDF network in each case and only change the meshing procedure. For BP, we decompose the time into sampling and meshing times.	}
	\label{tab:meshing_comparison}
	\begin{small}
	\begin{center}
\setlength{\tabcolsep}{3pt}
\begin{tabular}{cccc|ccc}
	\multicolumn{1}{l}{}      & \multicolumn{3}{c}{\textbf{Garments, $\phi_{\theta}$ network}}  & \multicolumn{3}{c}{\textbf{Cars, NDF network~\cite{Chibane20b}}} \\
	\multicolumn{1}{c|}{}     & \textit{BP} & \textit{Inflation} & \textit{Ours} & \textit{BP} & \textit{Inflation} & \textit{Ours} \\ \hline
	\multicolumn{1}{c|}{CHD ($\downarrow$)}  & 1.62          & 3.00              & \textbf{1.51}  & 6.84  & 11.24        & \textbf{6.63}     \\
	\multicolumn{1}{c|}{IC (\%, $\uparrow$)} & 92.51         & 88.48             & \textbf{92.80} & 90.50 & 87.09        & \textbf{90.87}    \\
	\multicolumn{1}{c|}{NC (\%, $\uparrow$)} & 89.50         & 94.16             & \textbf{95.50} & 61.50 & \textbf{73.19} & 70.38    \\
	\multicolumn{1}{c|}{Time ($\downarrow$)} & 16.5s + 3000s & \textbf{1.0 sec.} & 1.2 sec.       & 24.7s + 8400s  & \textbf{4.8 sec.}  & 7.1 sec.
	\end{tabular}
	\end{center}
	\end{small}
\end{table}

%% file: figs/choice_eps_AND_chd_mc_res.tex

\begin{figure}[t]
	\centering
		\begin{minipage}{.485\textwidth}
				\centering
		\includegraphics[clip, trim={0.36cm 0 1.1cm -0.5cm},width=.99\textwidth]{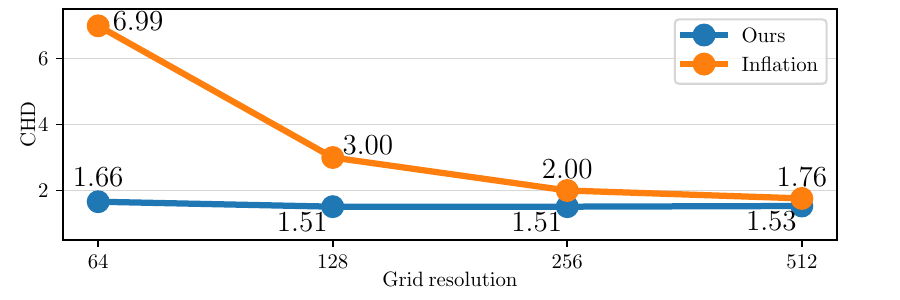}
		\captionof{figure}{\small \textbf{CHD as a function of grid resolution.} between reconstructed and ground truth meshes, averaged over the 300 training garments of MGN. \textit{Ours} yields constantly accurate meshes, \textit{Inflation} deforms the shapes at low resolutions.}
		\label{fig:CHD_MC_res}
	\end{minipage}
	\begin{minipage}{.485\textwidth}
		\centering
		\begin{overpic}[width=.97\textwidth]{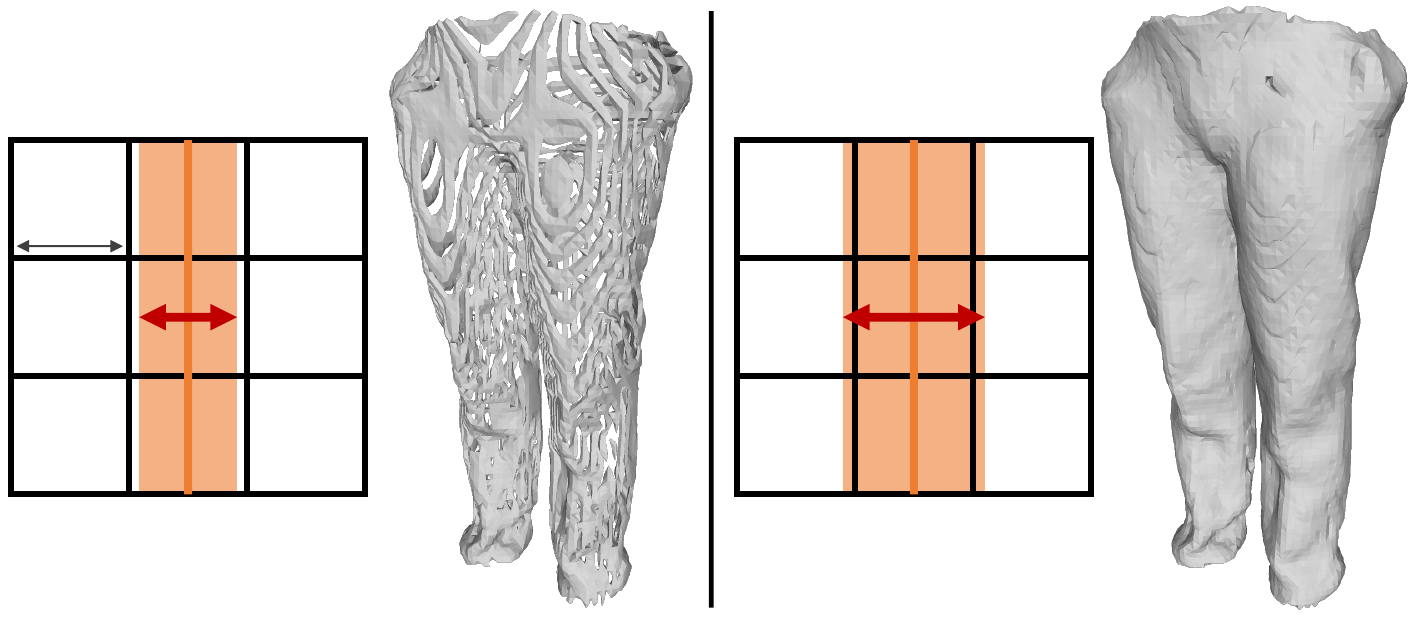}
			\put(5,28){\small{$s$}}
			\put(18,21){\small{\color{red}{$2\epsilon$}}}
			\put(69.7,21){\small{\color{red}{$2\epsilon$}}}
		\end{overpic}
		\captionof{figure}{\small \textbf{Choosing $\epsilon$ for \textit{Inflation}}. when meshing a UDF's $\epsilon$ iso-level with standard marching cubes, the value of $\epsilon$ is lower bounded by half the step size $s$. \textbf{Left}: $2\epsilon<s$ yields many large holes. \textbf{Right}: $2\epsilon \geq s$ yields a watertight mesh.}
		\label{fig:choice_eps}
	\end{minipage}%
\end{figure}

%% file: tables/optim_PC.tex

\begin{table}[t]
	\caption{\small \textbf{Fitting to sparse point clouds.} The table shows average Chamfer (CHD), image consistency (IC), and normal consistency (NC) wrt. ground truth test garments. We report metrics for un-optimized latent codes (\textit{Init.}), after optimizing ($\mathcal{L}_{PC,mesh}$) using our method, and optimizing either $\mathcal{L}_{PC,UDF}$ or $\widetilde{\mathcal{L}}_{PC,UDF}$ in the implicit domain.
	\textbf{(a)} A sparsely sampled ground truth mesh. \textbf{(b)} Mesh reconstructed by mimimizing $\mathcal{L}_{PC,mesh}$, \textbf{(c)} $\mathcal{L}_{PC,UDF}$, \textbf{(d)} $\widetilde{\mathcal{L}}_{PC,UDF}$.}
	\label{tab:optim_PC}
	\begin{small}
	\begin{center}
			\begin{tabular}{c|cccc}
				  & \textit{Init.} & $\mathcal{L}_{PC,mesh}$ & $\mathcal{L}_{PC,UDF}$ & $\widetilde{\mathcal{L}}_{PC,UDF}$ \\
          \hline
        CHD   ($\downarrow$) & 20.45 & \textbf{3.54} & 4.54 & 4.69\\
				IC   (\%,$\uparrow$) & 69.54 & \textbf{84.84} & 82.80 & 82.31\\
				NC   (\%,$\uparrow$) & 74.54 & \textbf{86.85} & 80.68 & 86.35
		\end{tabular}
		\begin{tabular}{cccc}
			(a) & (b) & (c) & (d) \\
			\includegraphics[width=.1\textwidth]{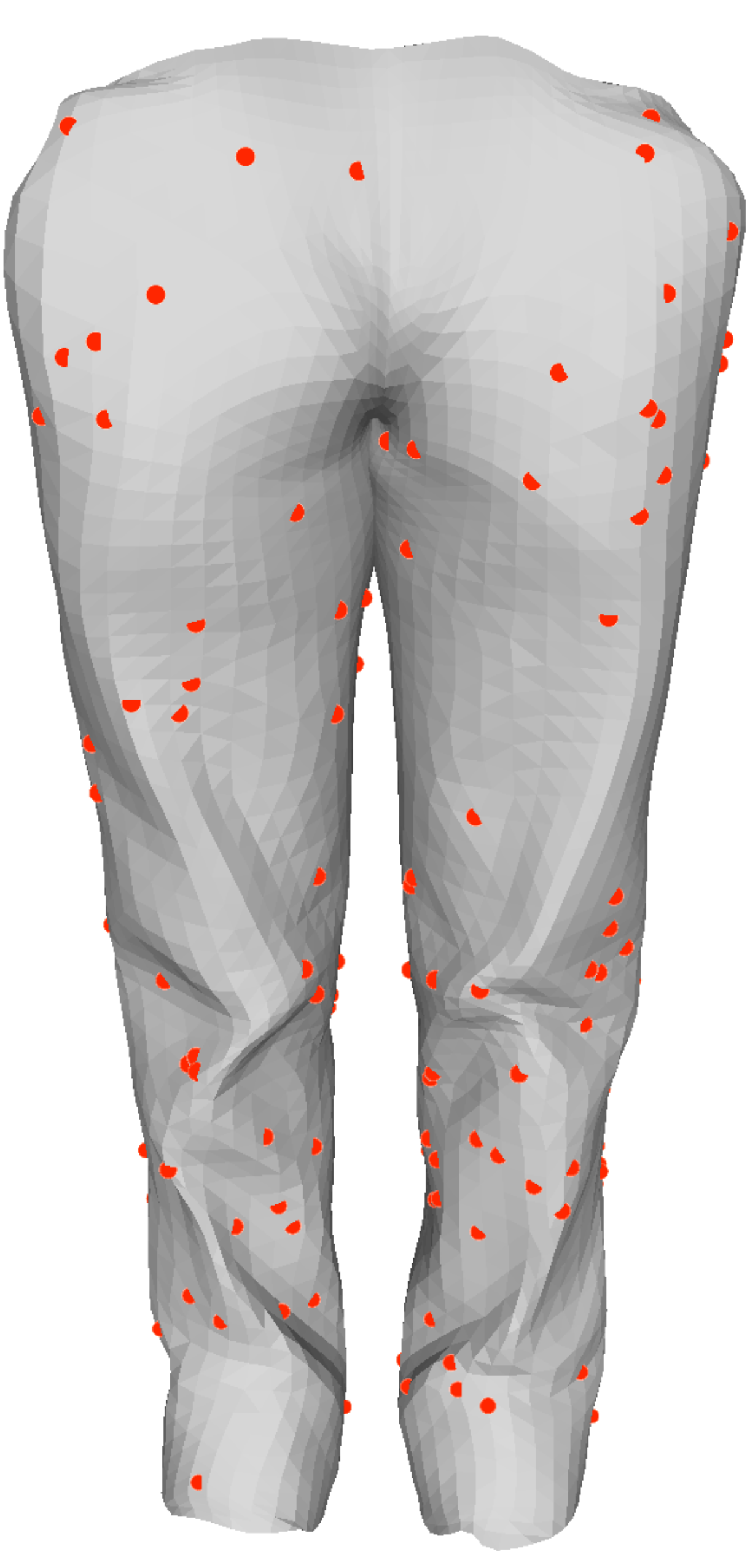}&
			\includegraphics[width=.1\textwidth]{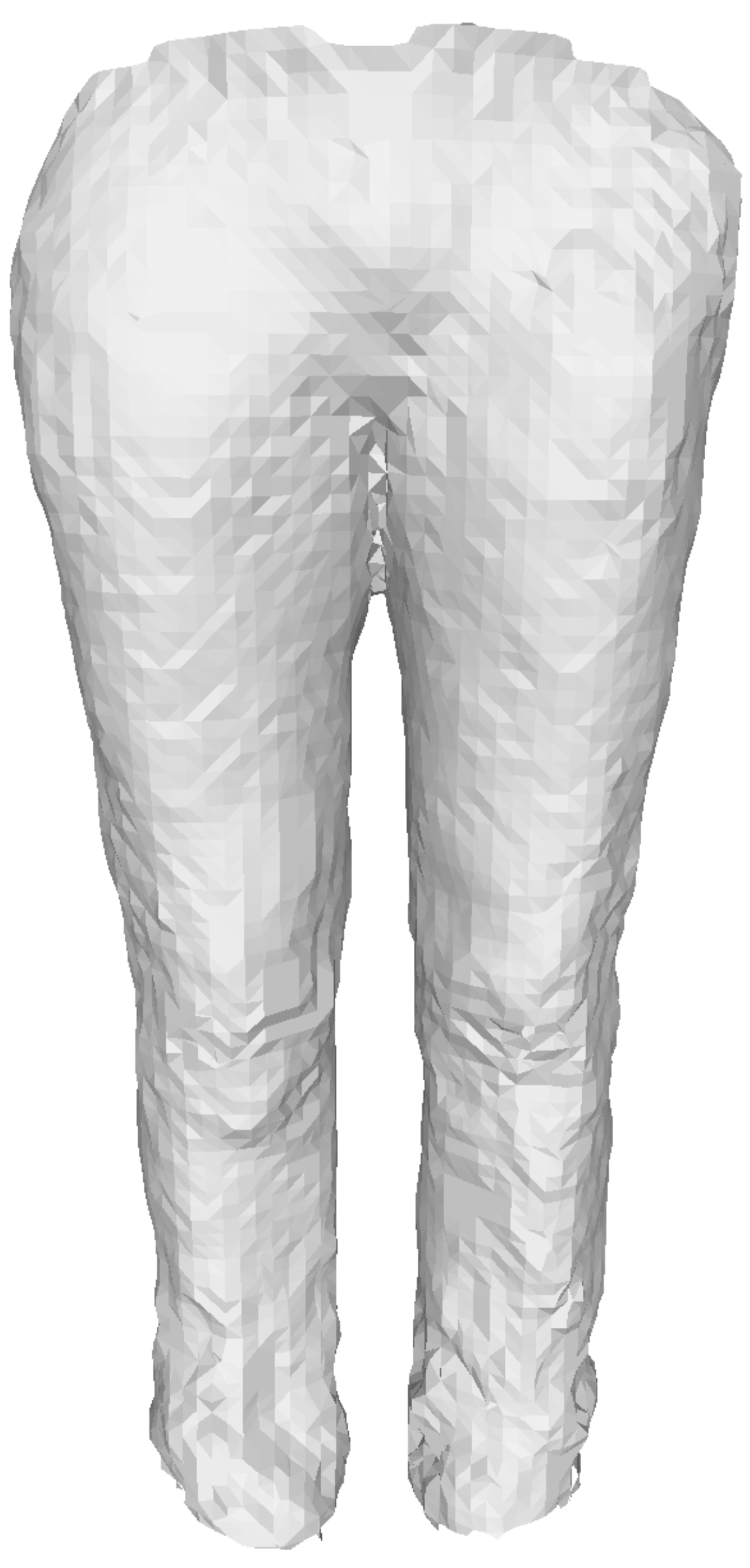} &
			\includegraphics[width=.1\textwidth]{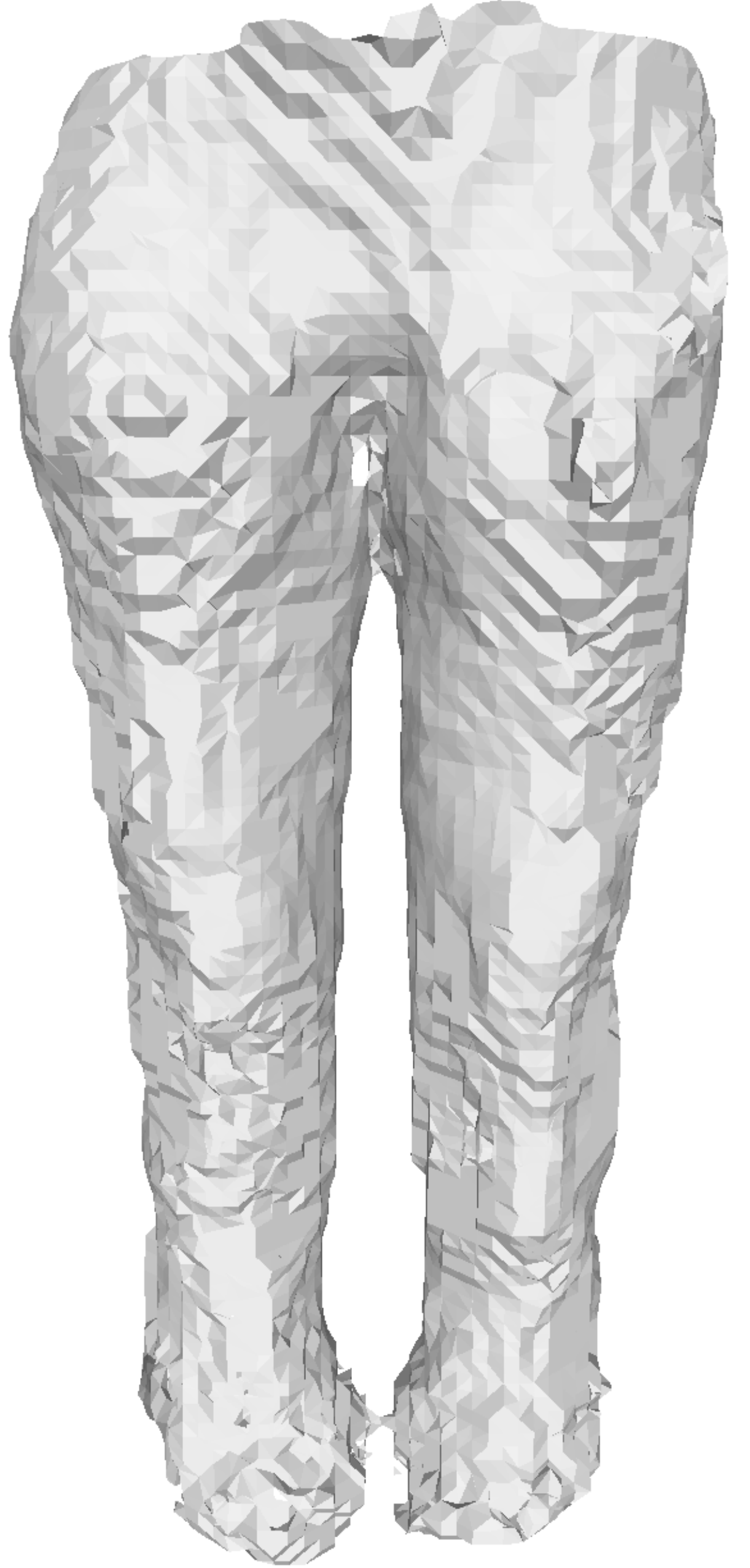} &
			\includegraphics[width=.1\textwidth]{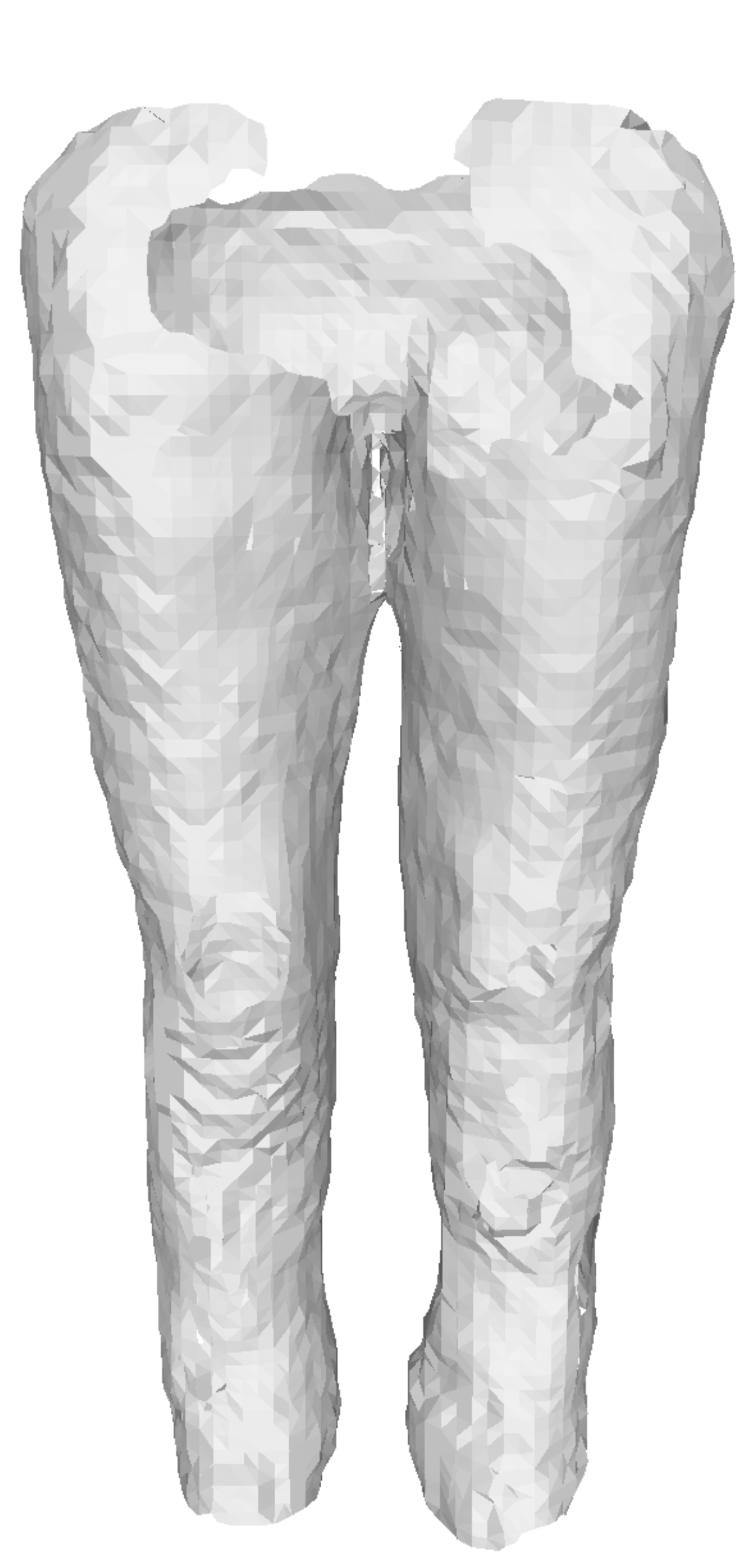}
		 \end{tabular}
	\end{center}
	\end{small}
\end{table}

%% file: tables/optim_silh.tex

\begin{table}[t]
	\caption{\small \textbf{Fitting to silhouettes.} Average Chamfer (CHD), image consistency (IC), and normal consistency (NC) wrt. ground truth test garments. We report metrics for un-optimized latent codes (\textit{Init.}), using our method to minimize ($\mathcal{L}_{silh,mesh}$), and by minimizing  ($\mathcal{L}_{silh,UDF}$) in the implicit domain. \textbf{(a)} Mesh reconstructed  by minimizing $\mathcal{L}_{silh,mesh}$. \textbf{(b,c)} Superposition of a target silhouette (light gray) and of the reconstructions (dark gray) by minimizing $\mathcal{L}_{silh,UDF}$ or $\mathcal{L}_{silh,mesh}$. Black denotes perfect alignment and shows that the $\mathcal{L}_{silh,UDF}$ mesh is much better aligned.}
	\label{tab:optim_silh}
	\begin{small}
	\begin{center}
			\begin{tabular}{c|ccc}
				  & \textit{Init.} & $\mathcal{L}_{silh,mesh}$ & $\mathcal{L}_{silh,UDF}$ \\
          \hline
        CHD    & 20.45 & \textbf{9.68} & 12.74 \\
				IC     & 69.54 & \textbf{79.90} & 74.46 \\
				NC     & 74.54 & \textbf{81.37} & 80.70 \\
		\end{tabular}
		\setlength{\tabcolsep}{-0pt}
		\begin{tabular}{ccc}
			(a) & (b) & (c) \\
			\includegraphics[width=.2\textwidth]{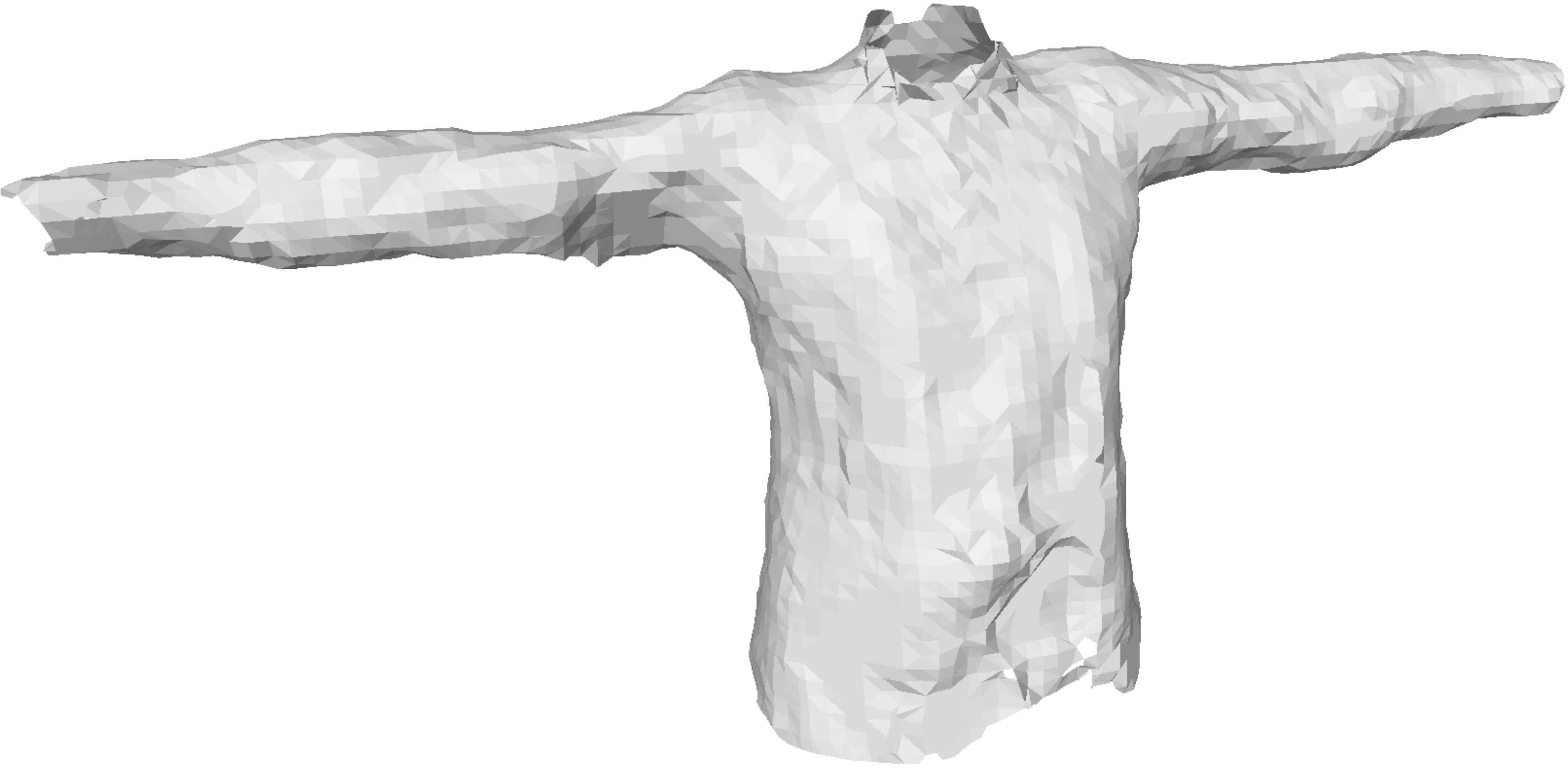}&
			\includegraphics[width=.2\textwidth]{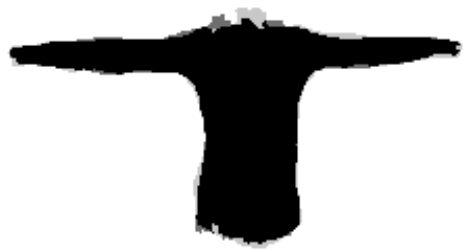} &
			\includegraphics[width=.2\textwidth]{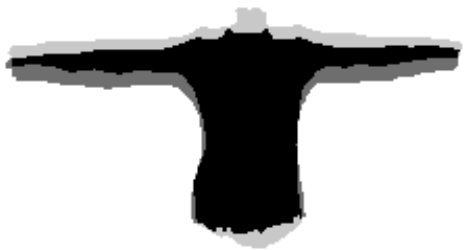}
		 \end{tabular}
		 \setlength{\tabcolsep}{6pt}
	\end{center}
	\end{small}
\end{table}

%% file: tables/ablate_grads.tex

\begin{table}[t]
	\begin{center}
	\caption{\small \textbf{Ablation Study.} Average Chamfer (CHD), image consistency (IC), and normal consistency (NC) for test garments using either our full approach to computing gradients (\textit{normals + border}) vs. computing the gradients everywhere using only the formula of Eq.~\ref{eq:derivatives0}. (\textit{normals}).}
	\label{tab:ablate_grads}
			\begin{tabular}{c|c|cc}
				Fitting & Metric & \shortstack{Gradients:\\ \textit{normals}} & \shortstack{Gradients:\\ \textit{normals} + \textit{border}}\\
				\hline
				\multirow{3}{*}{\shortstack{Point cloud, \\ $\mathcal{L}_{PC,mesh}$}}
				& CHD  	& 3.75  & \textbf{3.54}\\
				& NC  	& 84.28 & \textbf{84.84}\\
				& IC 		& 86.71 & \textbf{86.76}\\
				\hline

				\multirow{3}{*}{\shortstack{Silhouette, \\ $\mathcal{L}_{silh,mesh}$}}
				& CHD 	& 10.45 & \textbf{9.68}\\
				& IC 		& 78.84 & \textbf{79.90}\\
				& NC 		& 80.86 & \textbf{81.37}\\

		\end{tabular}
	\end{center}

\end{table}

%% file: figs/topology_change.tex

\setlength{\tabcolsep}{3pt}
\begin{figure*}[t]
	\begin{center}
	\begin{overpic}[width=1\textwidth]{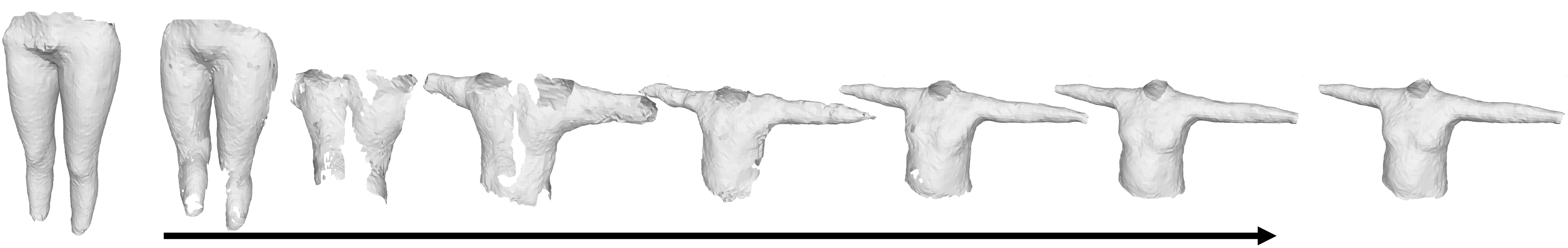}

		\put(3,-2){\small{(a)}}
		\put(44,-2){\small{(b)}}
		\put(90,-2){\small{(c)}}

	\end{overpic}
	\end{center}
		 \caption{\small \textbf{Optimization with a change in topology:} \textbf{(a)} Starting mesh associated to the initial latent code $\mathbf{z} = \mathbf{z}_{start}$ ; \textbf{(b)} Optimizing $\mathbf{z}$ with gradient descent by applying a 3D Chamfer loss between the reconstructed mesh and a target shape shown in \textbf{(c)}. During optimization, the latent code takes values that do not correspond to valid garments, hence the tears in our triangulations. Nevertheless, it eventually converges to the desired shape.}
	\label{fig:topology_change}
\end{figure*}
\setlength{\tabcolsep}{6pt}

%% file: figs/ndf_anchorudf.tex

\begin{figure}[t]
	\begin{center}
	 \includegraphics[width=\textwidth]{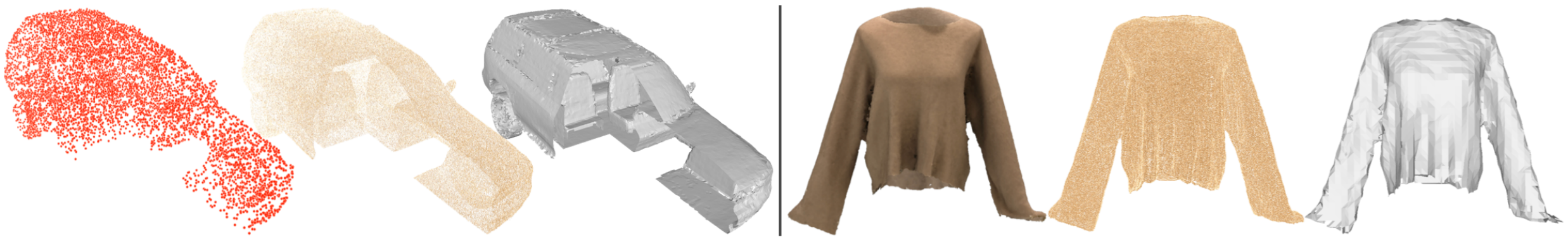} \\
	\end{center}
		\label{thisisatest}
		\caption{\small \textbf{Using our approach to triangulate the outputs of NDF~\cite{Chibane20b} (left) and AnchorUDF~\cite{Zhao21a} (right).} In both cases, we display the input to the network, a point cloud in one case and a color image in the other, the dense cloud of points that is the final output of these methods, and a triangulation of the UDF they compute generated using our method.}
		 \label{fig:ndf_anchorudf}
\end{figure}

%% file: tex/5_conclusion.tex

\section{Conclusion}
\label{sec:conclusion}

We have shown that deep-implicit non-watertight surfaces expressed in terms of unsigned distance functions could be effectively and differentiably triangulated. This provides an explicit parameterization of such surfaces that can be integrated in end-to-end differentiable pipelines, while retaining all the strengths of implicit representations, mainly that a network can accurately represent shapes with different topologies (jeans, sweater...) from the same latent space. In future work, we will explore how it can be used to jointly optimize the pose and clothes of people wearing loose attire.

%% file: tex/6_supp.tex

\section*{Supplementary Material}
\label{sec:supp}

In Sec.~\ref{sec:supp_training} we describe the procedure used to train the UDF network $\phi_{\theta}$ on MGN garments. In Sec.~\ref{sec:supp_metrics} we explain the metrics used in the experiment section. In Sec.~\ref{sec:supp_gradients} we explain in more details the gradients introduced in the main paper. In Sec.~\ref{sec:supp_approx_UDF} we show experimental evidence that artifacts appearing at high resolution are caused by the UDF field being approximated. In Sec.~\ref{sec:supp_voting} we demonstrate the benefits of the voting scheme for establishing pseudo-signs. In Sec.~\ref{sec:supp_optim_random_z}, we show the fitting to sparse pointclouds of Sec.~\ref{subsubsec:differentiability_demos_PC} can be initialized from random latent codes.

\subsection{Network Training}
\label{sec:supp_training}
We train one auto-decoder~\cite{Park19c} network $\phi_{\theta}$ to approximate the UDF field of a garment collection. We use the dataset from MGN~\cite{Bhatnagar19} consisting of 328 meshes from which we keep 300 instances for training and 28 for testing.

To generate UDF supervision sample points and values, meshes are scaled to fit a sphere of radius $0.8$, and for each mesh $i$ we generate $N$ training samples $(\mathbf{p}_{i,j}, d_{i,j}) \in \mathbb{R}^3\times \mathbb{R}^+$ where $d_{i,j}$ is the minimum between $d_{max}$ and the distance from 3D point $\mathbf{p}_{i,j}$ to the $i$-th shape. We clamp UDF values at $d_{max}=0.1$ to avoid wasting the network's capacity on learning a precise field away from the surface as in~\cite{Park19c,Chibane20b}. We pick $N=30000$ and sample $6000$ points uniformly on the surface, $12000$ within a distance $0.05$, $8000$ within a distance $0.3$ and $4000$ within the bounding box of side length $2$. As opposed to SDF values, unsigned distances $d_{i,j}$ can be computed directly from raw triangle soups with standard software~\cite{trimesh}, and do not require any pre-processing of the meshes.

$\phi_{\theta}$ is implemented by a 9-layer MLP with 512 hidden dimensions and ReLU activation functions. It uses Fourier positional encoding of fifth order on the 3D coordinate inputs~\cite{Sitzmann20}. We jointly optimize the network's weights $\theta$ with one latent vector embedding $\mathbf{z}_i \in \mathbb{R}^{128}$ per training shape $i$ by minimizing the $L_1$ loss between the predicted and target UDF values, with a regularization of strength $\lambda=10^{-4}$ on the norm of the latent codes. With $\mathcal{T}$ as the training set, the full loss is
$$
\mathcal{L} = \frac{1}{|\mathcal{T}|\cdot N} \sum_{i \in \mathcal{T}} \left [ \sum_{j=1}^N \left | \phi_{\theta}(\mathbf{z}_i, \mathbf{p}_{i,j}) - d_{i,j} \right |+ \lambda \left \| \mathbf{z}_i \right \|_2 \right ]
$$
and is minimized using Adam~\cite{Kingma14a} for 2000 epochs.

\subsection{Metrics}
\label{sec:supp_metrics}

Given a reconstructed mesh $\widetilde{M}$ and a set $A$ of points on its surface, along with a ground-truth mesh $M$ and a set $B$ of points on its surface, we %
\begin{itemize}
  \item \textbf{Chamfer distance.} We take it to be
 \begin{small}
 \begin{align}
  \mathrm{CHD}(\widetilde{M},M)
  &=   \frac{1}{|A|} \sum_{a \in A} \! \min_{b \in B} \left \| a - b \right \|^2  \\
  &+   \frac{1}{|B|} \sum_{b \in B}  \! \min_{a \in A} \left \| a - b \right \|^2  , \nonumber
  \end{align}
  \end{small}
the sum of the average distance of each point in $A$ to $B$ and the average distance of each point in $B$ to $A$.

  \item \textbf{Image consistency.} Let $K$ be a set of 8 cameras located at the vertices of a cuboid encompassing the garments looking at its centroid. For each $k\in K$ we render the corresponding binary silhouette $S_k \in \{0,1\}^{256 \times 256}$ (respectively $\widetilde{S}_k$) and normal map $N_k \in \mathbb{R}^{256 \times 256 \times 3}$ (resp. $\widetilde{N}_k$) of mesh $M$ (resp. $\widetilde{M}$). Then we define the image consistency between $\widetilde{M}$ and $M$ as
 \begin{small}
 \begin{align}
\!\!\!  \mathrm{IC}(\widetilde{M},M) = \tfrac{1}{|K|} \! \sum_{k \in K} \! \mathrm{IoU}(\widetilde{S}_k, S_k) * \mathrm{COS}(\widetilde{N}_k, N_k) \; ,
 \end{align}
 \end{small}
where $\mathrm{IoU}$ is the intersection-over-union of two binary silhouettes and $\mathrm{COS}$ is the average cosine-similarity between two normal maps. Both can be written as
\begin{small}
  \begin{align}
 \!\!\!  \mathrm{IoU}(\widetilde{S}, S) =& \! \sum_{u=1}^{H} \! \sum_{v=1}^{W} \! \widetilde{S}_{u,v} S_{u,v} \! \\
 & \cdot \left [\sum_{u=1}^{H} \! \sum_{v=1}^{W} \! max(\widetilde{S}_{u,v} + S_{u,v},1)  \right ]^{-1} \; ,
  \end{align}
  \end{small}
\begin{small}
  \begin{align}
 \!\!\!  \mathrm{COS}(\widetilde{N}, N) = \tfrac{1}{HW} \! \sum_{u=1}^{H} \! \sum_{v=1}^{W} \! \tfrac{\widetilde{N}_{u,v} \cdot N_{u,v}}{\left \| \widetilde{N}_{u,v} \right \| \left \| N_{u,v} \right \| } \; , \nonumber
  \end{align}
  \end{small}
with $H$ and $W$ the image height and width, $S_{u,v} \in \{0, 1\}$ the binary pixel value at coordinate $(u,v)$ of $S$ and $N_{u,v} \in \mathbb{R}^3$ the color pixel value at coordinate $(u,v)$ of $N$.

  \item \textbf{Normal consistency.} We take it to be
 \begin{small}
\begin{align}
\!\!\! \mathrm{NC}(\widetilde{M},M) &
	\! = \!\tfrac{1}{|A|}\!\! \sum_{a \in A} \!\! \left |\cos[ \mathbf{\widetilde{n}}(a), \mathbf{n}(\underset{b \in B}{\arg\!\min} \left \| a - b \right \|^2 )  ]\right| \nonumber \\
	 + & \tfrac{1}{|B|}\!\! \sum_{b \in B} \!\! \left   |\cos [ \mathbf{n}(b), \mathbf{\widetilde{n}}(\underset{a \in A}{\arg\!\min} \left \| a - b \right \|^2 )  ]\right|,
\end{align}
 \end{small}
the average \textit{unsigned} cosine-similarity between the normals of pairs of closest point in $A$ and $B$, where  $\mathbf{\widetilde{n}}(x)$ denotes the normal at point $x$.

\end{itemize}

\subsection{Differentiating through Iso-Surface Extraction}
\label{sec:supp_gradients}

In Sec.~\ref{sec:differentiating}, we derived gradients for surface points with respect to the latent code $\bz$. We here expand on the underlying assumptions and justify our choices.

\parag{Using the $\alpha$-isolevel.}
\input{figs/proof_diff.tex}
Let $\alpha > 0$ be a small scalar and $\bz \in \mathbb{R}^C$ a latent code parametrizing the UDF field $\phi(\bz, \cdot)$. We consider $\bv \in \mathbb{R}^3$, a surface point lying within a facet of the mesh $M_{\bz}$, and $\bn$ its surface normal --defined up to its orientation. $\bv$ lies on the $0$-levelset of the field, and we choose to formulate it as the following linear combination
\begin{align}
  \bv = \tfrac{1}{2} (\bv_- + \bv_+) \; , \label{eq:mapping_v}
\end{align}
where
\begin{align}
  \bv_+ = \bv + \alpha \bn  \; \text{ and } \;  \bv_- = \bv - \alpha \bn \; . \nonumber
\end{align}
The arrangement of $\bv$, $\bv_+$ and $\bv_-$ is depicted in Fig.~\ref{fig:proof_diff}.

Both $\bv_+$ and $\bv_-$ are at a distance $\alpha$ from $\bv$. Assuming such points to belong to the $\alpha$-levelset, the outwards pointing normal of $\bv_+$ on the $\alpha$-levelset is $\bn$, and the one of $\bv_-$ is $-\bn$, and we can use~\cite{Atzmon19,Remelli20b} to write
\begin{align}
  \frac{\partial \bv_+}{\partial \bz} = - \bn \frac{\partial \phi}{\partial \bz}(\bz, \bv_+) \; \; \; \; \; \; \text{ and } \; \; \; \; \; \; \; \frac{\partial \bv_-}{\partial \bz} =  \bn \frac{\partial \phi}{\partial \bz}(\bz, \bv_-)\; .
  \label{eq:equal_derivatives_vPlusMinus}
\end{align}
and differentiating the mapping of Eq.~\ref{eq:mapping_v} --which we consider as fixed-- yields
\begin{align}
  \frac{\partial \bv}{\partial \bz} = \frac{\bn}{2} \left [ \frac{\partial \phi}{\partial \bz}(\bz, \bv - \alpha \bn) - \frac{\partial \phi}{\partial \bz}(\bz, \bv + \alpha \bn)\right ]   \; . \label{eq:equal_derivatives0}
\end{align}

\parag{Approximate gradients.}

In practice however, $\bv_+$ and $\bv_-$ are not guaranteed to lie on the $\alpha$-levelset, but can be on a $\beta$-levelset with $\beta < \alpha$, in which case their normals differ from $\bn$ and $-\bn$. For our assumption to hold, $\bv$ needs to be the closest point to $\bv_+$ on the $0$-levelset, and similarly for $\bv_-$, which is true when $\alpha$ is small compared to the surface curvature. We thus use Eqs.~\ref{eq:equal_derivatives_vPlusMinus}, \ref{eq:equal_derivatives0} as approximations only.

Eq.~\ref{eq:equal_derivatives0} is only flawed for points with high curvature, and still holds true for most of the points lying on unwrinkled regions of the surface. Since gradients backpropagated to the latent code are averaged over the entire surface (as in~\cite{Mueller22}), a minority of them being noisy is not an issue. Sec.~\ref{subsec:differentiability_demos} empirically shows that using $\alpha=0.01$ works in practice for a wide range of shapes.

\parag{Uniqueness of the mapping.}

Eq.~\ref{eq:mapping_v} is an arbitrary choice of a mapping. It is not unique, and one could instead pair $\bv$ to other points on the $\alpha$-levelset, leading to a different result in Eq.~\ref{eq:equal_derivatives0}. We deliberately chose the 2 closest points to naturally surround $\bv$ with its closest neighbors.

\parag{Minimizing a downstream loss.}

Eq.~\ref{eq:equal_derivatives0} can be used to minimize downstream loss functions directly defined on mesh vertices with gradient descent. Given such a loss function $\mathcal{L}$, we use the chain rule to write
\begin{align}
  \frac{\partial \mathcal{L}}{\partial \bz} = \sum_{(\bv, \bn) \in M_{\bz}} \frac{\partial \mathcal{L}}{\partial \bv} \frac{\bn}{2} \left [ \frac{\partial \phi}{\partial \bz}(\bz, \bv - \alpha \bn) - \frac{\partial \phi}{\partial \bz}(\bz, \bv + \alpha \bn)\right ]   \;\;\;\; . \nonumber
\end{align}

We rely on the field being an UDF and move its zero level set. This is in practice enforced by freezing the network weights, which is thus acting as a strong prior on the field, and only optimize the latent code.

\parag{The case of border points.}
\input{figs/proof_diff_border.tex}
Border do not only have 2 closest neighbors on the $\alpha$-levelset, but an entire semi-circle as depicted in blue on Fig.~\ref{fig:proof_diff_border}. In this case, we pair $\bv$ with the outmost point on the $\alpha$-level set with
\begin{align}
  \bv = \bv_o - \alpha \bo \; , \label{eq:mapping_v_ext}
\end{align}
and follow the same reasoning as above. We consider $\bo$ as a mapping direction, and thus locally fixed.

\input{figs/outwards_vec_border.tex}

\parag{Constructing the $\bo$ vectors.} Fig.~\ref{fig:outwards_vec_border} depicts the outwards pointing vectors $\bo$ for one reconstructed garment. They are computed as follows. Let $\bv$ be a vertex lying on the border, $\bn$ be the normal vector of the facet it belongs to, and $\mathbf{e}$ be the border edge it is on.
We take $\bo$ to be
\begin{equation}
\bo= \omega  \frac{\bn \times \mathbf{e}}{\left \| \bn \times \mathbf{e} \right \|}  \;\;\;\; \text{with} \;\;\;\; \omega=\pm 1\; ,
\end{equation}
the unit vector colinear to the cross product of $\bn$ and $\mathbf{e}$.
This way, $\bo$ is both in the tangent plane of the surface and perpendicular to the border. We choose the sign $\omega$ to orient $\bo$ outwards. We write
\begin{equation}
\omega = \underset{\{-1,1\}}{\arg\!\max }  \;\; u(\bv+\omega \frac{\bn \times \mathbf{e}}{\left \| \bn \times \mathbf{e} \right \|} ) \; ,
\end{equation}
that is, we evaluate the UDF in both directions and pick the one that yields the highest value.

\subsection{Meshing approximate or real UDFs}
\label{sec:supp_approx_UDF}
\input{figs/mesh_real_udf.tex}

In Sec.~\ref{subsec:limitations} and Fig.~\ref{fig:CHD_MC_res} of the main paper, we mention artifacts of our meshing procedure when applied to approximate UDFs and at a high resolution. This is depicted in Fig.~\ref{fig:mesh_real_udf}\textbf{(b)}, where meshing a UDF represented by a shallow network (4 layers) with a grid resolution of 512 yields a mesh that is not smooth.

We hypothesized that this is due to the $0$-levelset of the field being slightly inflated into a volume, with many grid locations evaluating to a $0$ distance near the surface. This impedes Marching Cube's interpolation step and produces this staircase artifact. To validate this hypothesis, in Fig.~\ref{fig:mesh_real_udf}\textbf{(c)} we apply our meshing procedure to the exact UDF grid, numerically computed from the ground truth mesh of Fig.~\ref{fig:mesh_real_udf}\textbf{(a)}. This results in a smooth surface, thus indicating that the staircase artifact is indeed a consequence of meshing approximate UDFs.

\subsection{Ablation study: pseudo-sign and breadth-first exploration}
\label{sec:supp_voting}
\input{figs/method.tex}

In Sec.~\ref{sec:method} we described a way to locally compute the pseudo-signed distance using gradient orientations (\textit{PSD}), that is described in more details in Fig.~\ref{fig:method_pseudosign}. The \textit{PSD} method has two shortcomings. First, the choice of the anchor corner implies that the anchor will have a positive pseudo-sign, and thus choosing a different anchor might invert all the signs of the cell. Since the choice is arbitrary, adjacent cells might have opposing sign choices: they will produce meaningful facets, but with opposing orientations. This can be partially fixed in a post-processing step that scans the mesh trying to consistently reorient the facets, but this proved to be a time-consuming operation and it does not always find a consistent orientation.
Second, if the surface in the cell or in the immediate proximity is not smooth enough, the gradients of the field can have ambiguous orientations (i.e. they do not clearly oppose each other, for example at a $45^{\circ}$ angle). In this setting, two different anchors can produce different pseudo-signs for the corners of the cell, and thus nearby cells that use a different anchor can assign different pseudo-signs to the same corner. This inconsistency creates an unwanted hole in the mesh and happens especially with learned UDF fields, which have noisy gradients.

The breadth-first exploration (\textit{BFE}) method with a voting scheme that we propose has the purpose of improving these shortcomings: produce consistent normal orientations in adjacent facets and increase the robustness of the method on learned UDF fields with noisy gradients. The first objective is reached thanks to the breadth-first exploration itself, which is implemented using queues: following the surface makes it possible to store values of previously computed pseudo-signs, ensuring that corners have the same pseudo-sign in adjacent cells. This also reduces the number of dot products required to complete the meshing procedure, since corners are only computed once instead of being recomputed in every cell. However, simply plugging the pseudo-signed distance computation in this breadth-first exploration can cause even more artifacts due to anchor choice, as they can propagate in nearby cells since the cells are not treated independently anymore.

To solve this problem and at the same time address the second objective, we use the voting scheme described in Sec.~\ref{sec:method}. This voting scheme has been experimentally inferred by looking at artifacts of the previous procedure, and has three motivations. First, it avoids an explicit and arbirary anchor choice, which is the main cause of inconsistencies, and it increases the robustness by making multiple neighbours vote for a single corner. Second, it prevents votes to be computed along diagonals in a cell, because the underlying interpolation algorithm of marching cubes does not create vertices along cell diagonals. Third, it prevents gradients facing each other along an edge to vote for having an opposing sign --when they indicate a local maximum of the field instead.

Moreover, we notice that in corners with possible ambiguities the absolute value of the sum of received votes will be low. Some neighbours will vote positively and some others negatively, and the weight itself of the votes can be low when gradients are not clearly facing or opposing each other. We detect these cases that get a sum of votes below a threshold, fixed to $\cos(\pi/4)$, and we put them into a separate queue with a lower priority, to be re-evaluated later. The threshold has been set by noticing that, in a single-vote scenario, gradients at a $[45^{\circ},135^{\circ}]$ angle have a high ambiguity, since a $45^{\circ}$ variation in the angle would flip the sign of their dot product. This queue is explored when the main exploration is over, thus increasing the number of neighbours that can vote and making the sign decision more robust.
We also employ a third queue, which is explored with the lowest priority, that contains cells with multiple non-adjacent facets. These cells can potentially start the exploration of a non-contiguous surface, and are thus explored at the very end.

\input{figs/orientations_holes.tex}
\input{tables/ablate_BFE.tex}

To validate this algorithm, we compare \textit{BFE} with the simple application of \textit{PSD} using the same garment network and dataset described in Sec.~\ref{subsec:accuracy_speed} (and Tab.~\ref{tab:meshing_comparison} left), at different resolutions. The post-processing steps (Fig.~\ref{fig:artefact}) applied to the two methods are the same except for the parameters used. Since \textit{PSD} produces slightly less precise borders, we apply a coarser filtering of spurious facets and remove those whose UDF value is larger than $1/6$ of the side-length of a cubic cell instead of half. In \textit{PSD} we also apply 5 steps of laplacian smoothing on the borders instead of 1 for \textit{BFE}.

In Fig~\ref{fig:orientations_holes} we see that \textit{BFE} produces consistent facets orientations, while \textit{PSD} does not. Moreover, one can notice small holes in the garments reconstructed with \textit{PSD} (center row), which tend to increase in number and decrease in dimension as the resolution increases, whereas \textit{BFE} is able to close most of them (bottom row), proving to be more robust.
Tab.~\ref{tab:ablate_BFE} shows that the \textit{BFE} method produces meshes with a slightly lower Chamfer distance, except at resolution 64. Since the size of the holes produced by \textit{PSD} is very small, they do not significatively impact the CHD of this method. They however produce artifacts that are detrimental to the quality of the reconstructed mesh. To have a quantitative measure of this, given a ground-truth mesh $M$ and a reconstructed mesh $\widetilde{M}$, we define the \textit{number of excess holes} as:
\begin{equation}
EH(\widetilde{M},M) = | |\widetilde{H}| - |H| | \; ,
\end{equation}
where $H$ and $\widetilde{H}$ are the sets of holes of $M$ and $\widetilde{M}$, computed as closed loops of edges that belong to a single triangle. This amounts to computing the number of holes in excess that are in $\widetilde{M}$ compared to $M$, or viceversa.

Tab.~\ref{tab:ablate_BFE} shows that \textit{BFE} has a consistent advantage over \textit{PSD} in this metric across all tested resolutions. In both methods, the EH tends to increase with resolution, as the limits of the learned field are approached and the gradients become noisier. The same experiment with a network trained on only 4 garments yields better results on such garments, with the \textit{BFE} producing no excess holes at all 64-512 resolutions, and \textit{PSD} producing a similar amount to that shown in the table.

Finally, the \textit{BFE} method is also slightly faster than \textit{PSD}. This is mainly due to the reduced number of dot products computed. In \textit{PSD} we compute 8 dot products per cell --which amounts to an average of 4 dot products per corner, since every corner belongs to 4 different cells. In \textit{BFE} each corner receives votes from a maximum of 6 neighbours with existing pseudo-signs. Since the exploration starts from one cell and proceeds breadth-first, for the vast majority of corners only a smaller number of neighbours will actually vote, decreasing the total number of dot products.

\subsection{Optimization from random initial latent codes}
\label{sec:supp_optim_random_z}
\input{tables/optim_PC_random_code.tex}
In Tab.~\ref{tab:optim_PC_random_code} we reproduce the experiment from Sec.~\ref{subsubsec:differentiability_demos_PC} and fit latent codes using sparse point clouds, but start from random latent codes instead of codes from a similar semantic class. This shows that the latter is not even a requirement because, despite starting from much worse initializations, our approach still succeeds better than direct supervision on the UDF values. Starting with latent codes of the same object category remains a plausible scenario because such codes could be provided by a regressor.

\clearpage
\subsection{Additional results}
\input{figs/additional_results_ndf.tex}
\input{figs/anchorudf.tex}
\input{figs/additional_results_rebut.tex}

In Fig.~\ref{fig:additional_results_ndf} we show additional results of our method applied to mesh the UDF regressed from NDF~\cite{Chibane20b} from sparse input point clouds. In Fig.~\ref{fig:anchorudf}, we mesh the UDF predicted by AnchorUDF~\cite{Zhao21a} from input images. In Fig.~\ref{fig:additional_results_rebut}, we compare our meshing method to the $\epsilon$-inflation baseline for other garment samples reconstructed by an auto-decoder network.

%% file: figs/proof_diff.tex

\begin{figure}
	\begin{center}
	\begin{overpic}[width=.48\textwidth]{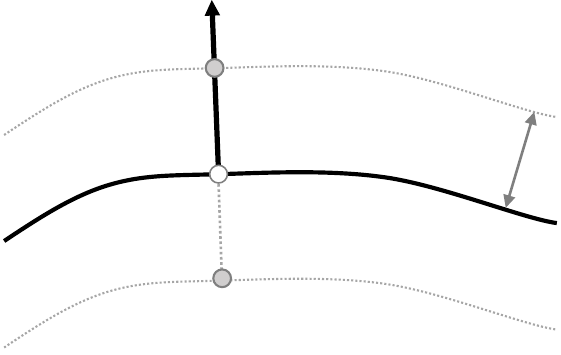}
		\put(40,60){\Large{$\bn$}}
		\put(40,53){\Large{$\bv_+$}}
		\put(41,27){\Large{$\bv$}}
		\put(41,8){\Large{$\bv_-$}}
		
		\put(94,34){\Large{$\alpha$}}

		\put(23,43){\textcolor{gray}{\Large{$S_{\alpha}$}}}
		\put(23,24){\Large{$S$}}
	\end{overpic}

  \vspace{-10pt}
	\end{center}
		 \caption{\textbf{Iso-surface differentiation}: $S$ is the minimum-levelset of UDF $\phi(\bz, \cdot)$ and $S_{\alpha}$ its $\alpha$-levelset ($\alpha > 0$). By using already established differentiability results on $\bv_+ \in S_{\alpha}$ and $\bv_- \in S_{\alpha}$, we derive new derivatives for $\bv \in S$.}
	\label{fig:proof_diff}
	\vspace{-5pt}
\end{figure}

%% file: figs/proof_diff_border.tex

\begin{figure}[t]
	\begin{center}
	\begin{overpic}[width=.47\textwidth]{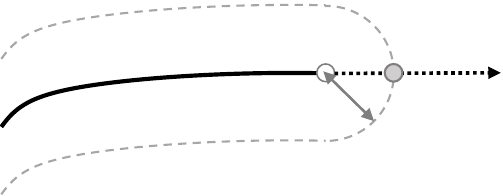}
		\put(95,19){\Large{$\bo$}}

		\put(58,26){\Large{$\bv$}}
		\put(80,27){\Large{$\bv_o$}}
		
		\put(64,16){\Large{$\alpha$}}

		\put(32,26){\Large{$S$}}
		\put(32,13){\textcolor{gray}{\Large{$S_{\alpha}$}}}
	\end{overpic}

  \vspace{-15pt}
	\end{center}
		 \caption{\textbf{Iso-surface differentiation at borders}: $S$ is the minimum-levelset of UDF $\phi(\bz, \cdot)$ and $S_{\alpha}$ its $\alpha$-levelset ($\alpha > 0$). By using already established differentiability results on $\bv_o \in S_{\alpha}$, we derive a new derivative for $\bv \in S$.}
	\label{fig:proof_diff_border}
	\vspace{-8pt}
\end{figure}

%% file: figs/outwards_vec_border.tex

\begin{figure}[t]
	\begin{center}
	\includegraphics[width=.6\textwidth]{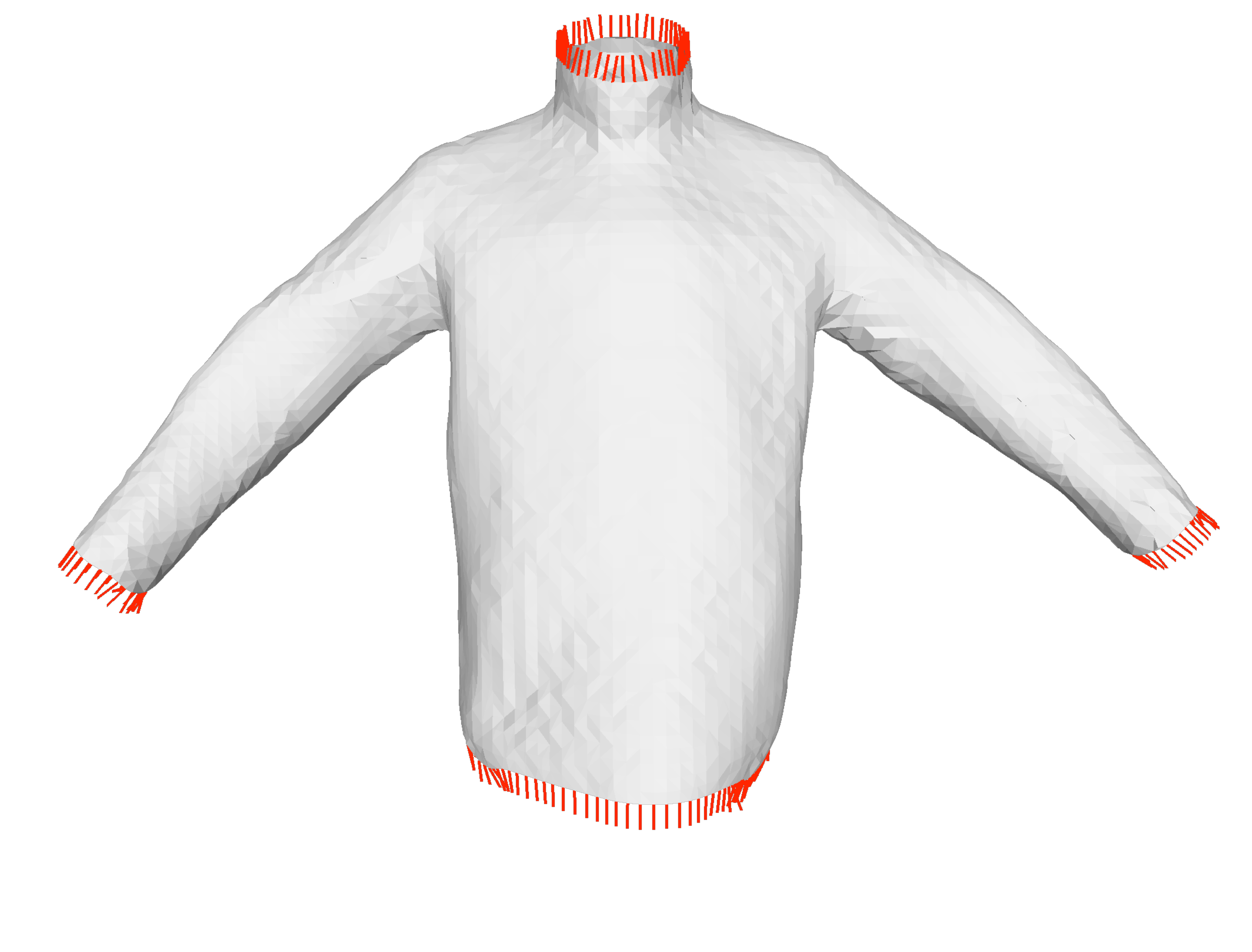}
  \vspace{-35pt}
	\end{center}
		 \caption{\textbf{Outwards pointing vectors}: for border vertices we define outwards pointing vectors $\bo$ to construct derivatives allowing the surface to shrink or extend along them.}
	\label{fig:outwards_vec_border}
	\vspace{-8pt}
\end{figure}

%% file: figs/mesh_real_udf.tex

\begin{figure}[t]
	\centering
	\begin{overpic}[width=.8\textwidth]{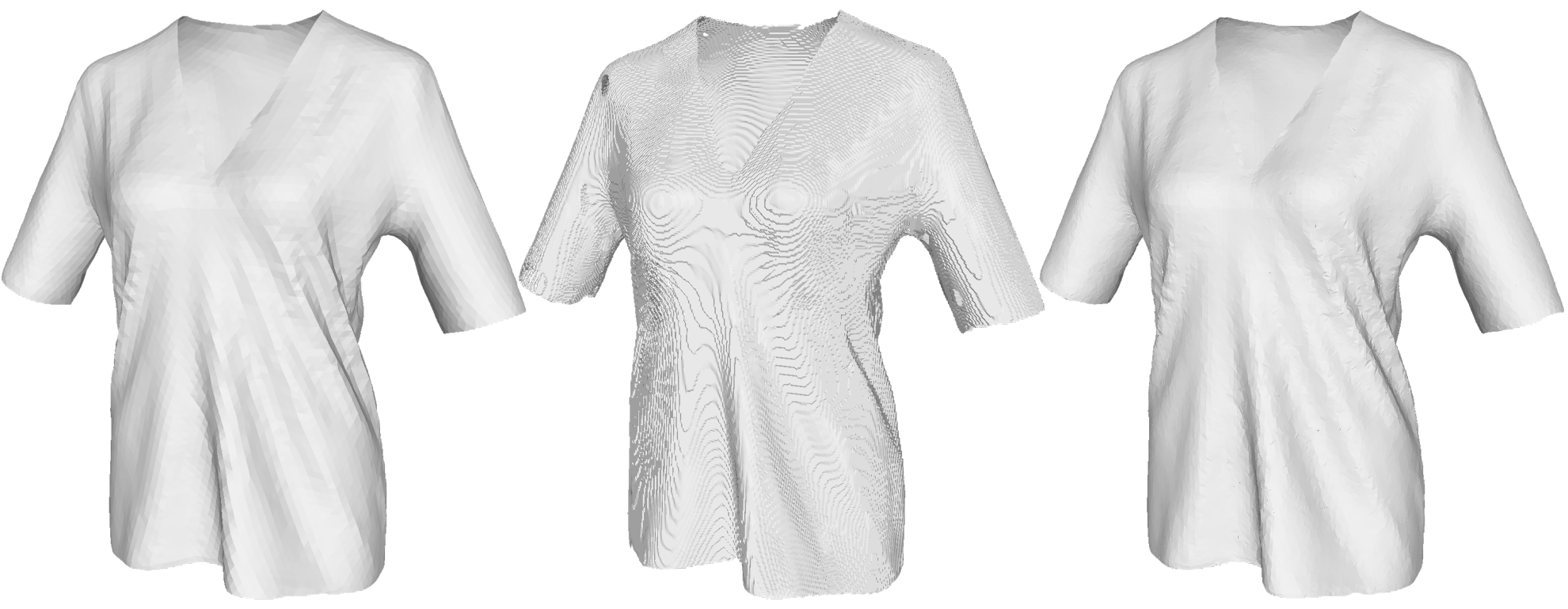}
		\put(16,-3){\small{(a)}}
		\put(49,-3){\small{(b)}}
		\put(82,-3){\small{(c)}}
	\end{overpic}
		 \caption{\textbf{Meshing UDFs:} \textbf{(a)} Ground truth mesh; \textbf{(b)} Our meshing procedure applied to a shallow UDF neural network yields staircase artifacts at a very high resolution (512); \textbf{(c)} Our method applied to the exact UDF at the same resolution reconstructs a smooth surface.}
	\label{fig:mesh_real_udf}
	\vspace{-10pt}
\end{figure}

%% file: figs/method.tex

\begin{figure}[t]
	\begin{center}
	\begin{overpic}[width=.6\textwidth]{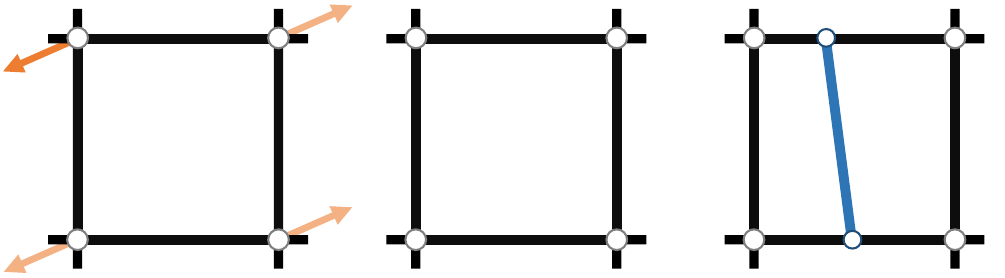}









		\put(8.5,20.5){\small{$u_1$}}
		\put(2,18.5){\small{$\bgrad_1$}}

		\put(22.5,20.5){\small{$u_2$}}
		\put(28.5,21){\small{$\bgrad_2$}}

		\put(22.5,5){\small{$u_3$}}
		\put(28.5,7){\small{$\bgrad_3$}}

		\put(8.5,5){\small{$u_4$}}
		\put(0,3){\small{$\bgrad_4$}}

		\put(42,25.){\small{$s_1 \text{=} u_1$}}

		\put(47,20.5){\small{$s_2 \text{=-} u_2$}}

		\put(47,.7){\small{$s_3 \text{=-} u_3$}}

		\put(42,5){\small{$s_4 \text{=} u_4$}}

		\put(83.5,19.5){\small{$\tfrac{s_1}{s_1-s_2}$}}

		\put(15,-4){\small{(a)}}
		\put(49,-4){\small{(b)}}
		\put(83,-4){\small{(c)}}

	\end{overpic}

  \vspace{-10pt}
	\end{center}
		 \caption{\textbf{Detecting surface crossings}: \textbf{(a)} all corners of the grid's cell are annotated with unsigned distance values $u_i$ and gradients $\bgrad_i$ ; \textbf{(b)} we locally approximate signed distances with $s_i {=} \text{sgn}(\bgrad_1 \cdot \bgrad_i) u_i$ ; \textbf{(c)} marching cubes processes these pseudo-signed distances and produces a surface element accordingly.}
	\label{fig:method_pseudosign}
	\vspace{-4mm}
\end{figure}

%% file: figs/orientations_holes.tex

\begin{figure}[t]
	\begin{center}
	\begin{overpic}[width=1.0\textwidth]{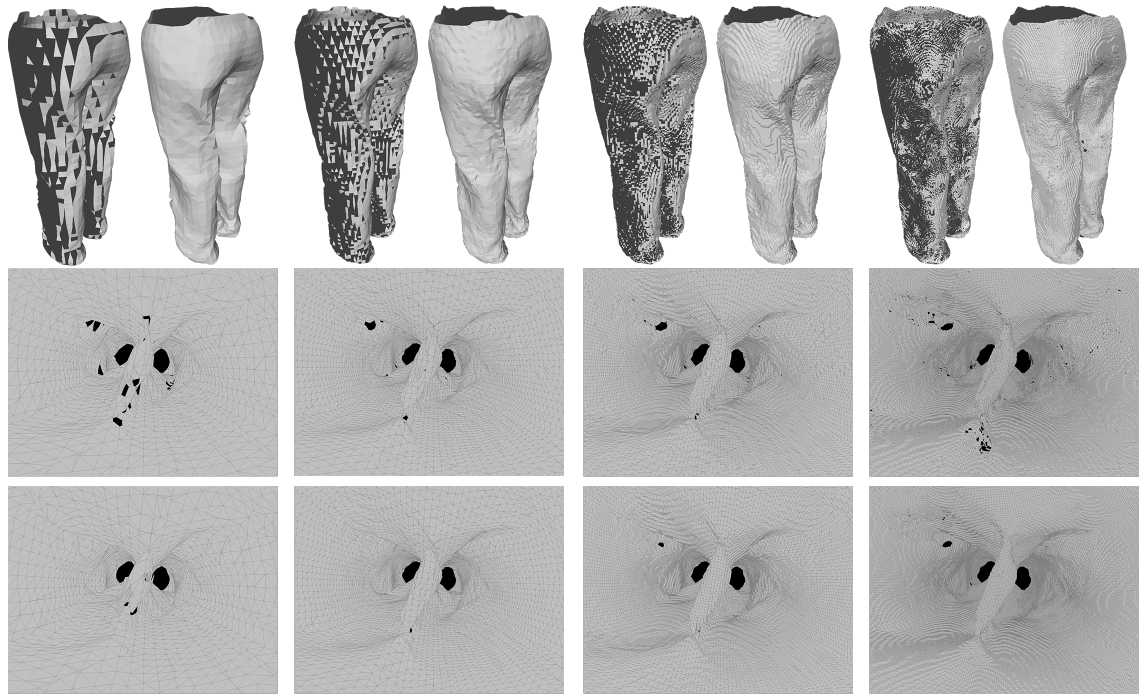}

		\put(9,-2){\small{(a) 64}} 
		\put(33,-2){\small{(b) 128}}
		\put(58,-2){\small{(c) 256}}
		\put(83,-2){\small{(d) 512}}

	\end{overpic}
  \vspace{-9pt}
	\end{center}
	
		 \caption{\small \textbf{Comparing qualitative results of PSD and BFE.} Each of the 4 columns corresponds to a meshing resolution, as indicated in the labels. In each column, top row left is the result of PSD, top row right is the result of BFE. Center and bottom rows show an above view of the same mesh, with holes colored in black. The two bigger holes correspond to the legs. Center row is PSD, bottom row is BFE.}
	\label{fig:orientations_holes}
	\vspace{-5mm}
\end{figure}
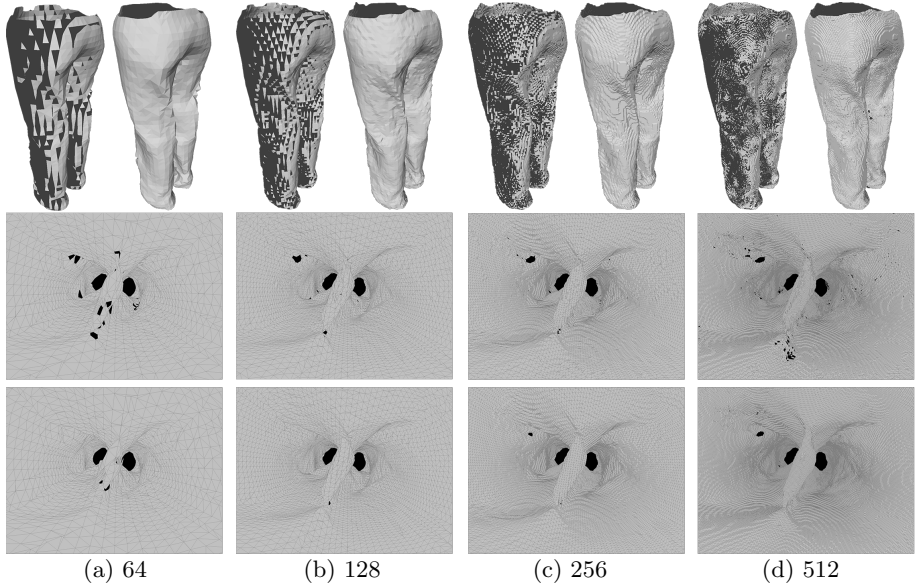

%% file: tables/ablate_BFE.tex

\begin{table}[t]
	\caption{\small \textbf{Comparing UDF meshing methods: pseudo sign (\textit{PSD}) versus breadth-first exploration with voting strategy (\textit{BFE}).} Average Chamfer distance (CHD), average number of excess holes (EH) and average processing time on 300 garments. We use a single UDF network and only change the meshing procedure.}
	\label{tab:ablate_BFE}
	\begin{small}
	\begin{center}
\setlength{\tabcolsep}{3pt}
\begin{tabular}{ccc|cc|cc|cc}
	\multicolumn{1}{c|}{Resolution}      & \multicolumn{2}{c|}{\textbf{64}}  & \multicolumn{2}{c|}{\textbf{128}} & \multicolumn{2}{c|}{\textbf{256}} & \multicolumn{2}{c}{\textbf{512}}\\
	\multicolumn{1}{c|}{Meshing procedure}     & \textit{PSD} & \textit{BFE} & \textit{PSD} & \textit{BFE} & \textit{PSD} & \textit{BFE} & \textit{PSD} & \textit{BFE} \\ \hline
	\multicolumn{1}{c|}{CHD ($\downarrow$)}  & \textbf{1.63}          & 1.66              & \textbf{1.51}  & \textbf{1.51}  & 1.52 & \textbf{1.51} & 1.61 & \textbf{1.53} \\
	\multicolumn{1}{c|}{EH ($\downarrow$)} & 21 & \textbf{1.6} & 153 & \textbf{7.8} & 1566 & \textbf{38} &  11526 & \textbf{478}   \\
	\multicolumn{1}{c|}{Time ($\downarrow$)} & 0.35s & \textbf{0.24s} & 1.4s & \textbf{1.2s} & 10.0s & \textbf{9.1s} & 105s & \textbf{69s}
	\end{tabular}
	\end{center}
	\end{small}
\end{table}

%% file: tables/optim_PC_random_code.tex

\begin{table}[t]
	\caption{\small \textbf{Fitting to sparse point clouds, with different latent code initializations:} either from a code of the same garment type (left), or from a random code (right). The table shows average Chamfer (CHD), image consistency (IC), and normal consistency (NC) wrt. ground truth test garments. We report metrics for un-optimized latent codes (\textit{Init.}), after optimizing ($\mathcal{L}_{PC,mesh}$) using our method, and optimizing either $\mathcal{L}_{PC,UDF}$ or $\widetilde{\mathcal{L}}_{PC,UDF}$ in the implicit domain.}
	\label{tab:optim_PC_random_code}
	\begin{small}
	\begin{center}
		\setlength{\tabcolsep}{1.5pt}
			\begin{tabular}{c|cccc|cccc}
				\multicolumn{1}{l}{}      & \multicolumn{4}{c}{\textbf{Initialization: same class}}  & \multicolumn{4}{c}{\textbf{Initialization: random}} \\
				  & \textit{Init.} & $\mathcal{L}_{PC,mesh}$ & $\mathcal{L}_{PC,UDF}$ & $\widetilde{\mathcal{L}}_{PC,UDF}$ & \textit{Init.} & $\mathcal{L}_{PC,mesh}$ & $\mathcal{L}_{PC,UDF}$ & $\widetilde{\mathcal{L}}_{PC,UDF}$ \\
          \hline
        CHD   ($\downarrow$) & 20.45 & \textbf{3.54} & 4.54 & 4.69			&129.51 & \textbf{3.64} & 4.59 & 4.60\\
				IC   (\%,$\uparrow$) & 69.54 & \textbf{84.84} & 82.80 & 82.31		& 49.08 & \textbf{84.70} & 83.22 & 82.94\\
				NC   (\%,$\uparrow$) & 74.54 & \textbf{86.85} & 80.68 & 86.35		& 56.74 & \textbf{86.96} & 84.20 & 86.62
		\end{tabular}
	\end{center}
	\end{small}
\end{table}

%% file: figs/additional_results_ndf.tex

\begin{figure}[t]
	\begin{center}
	 \includegraphics[width=\textwidth]{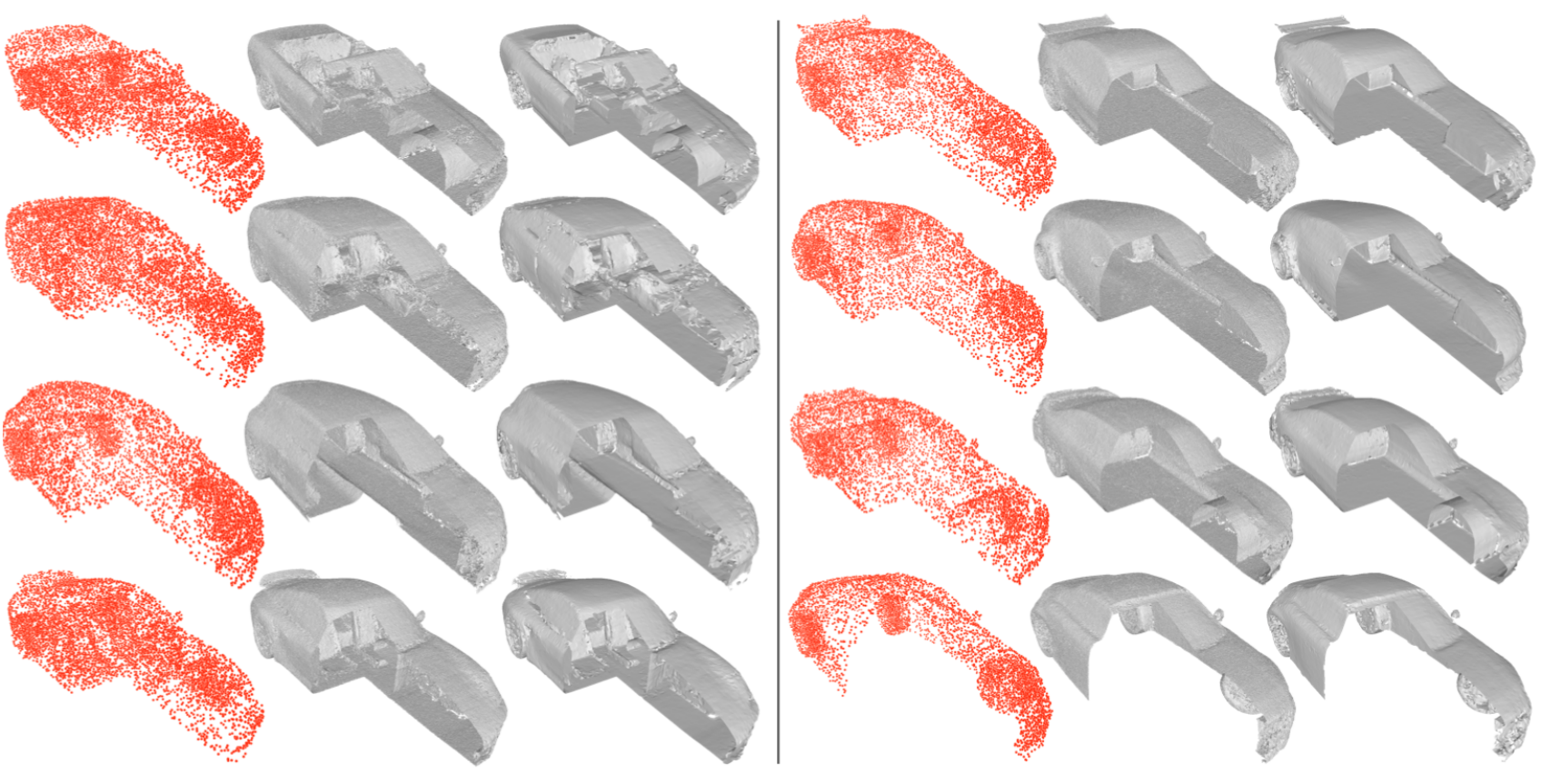} \\
	\end{center}
	\vspace{-5mm}
		\caption{\small \textbf{Using our approach to triangulate the outputs of NDF~\cite{Chibane20b}.} For 8 examples we display the input to the network (a sparse point cloud), a mesh of the predicted UDF mesh reconstructed by the ball pivoting method in more than 2 hours, and a triangulation of the UDF generated using our method in less than 10 seconds.}
		 \label{fig:additional_results_ndf}
	\vspace{-10pt}
\end{figure}

%% file: figs/anchorudf.tex

\begin{figure}
	\begin{center}
	 \includegraphics[width=.7\textwidth]{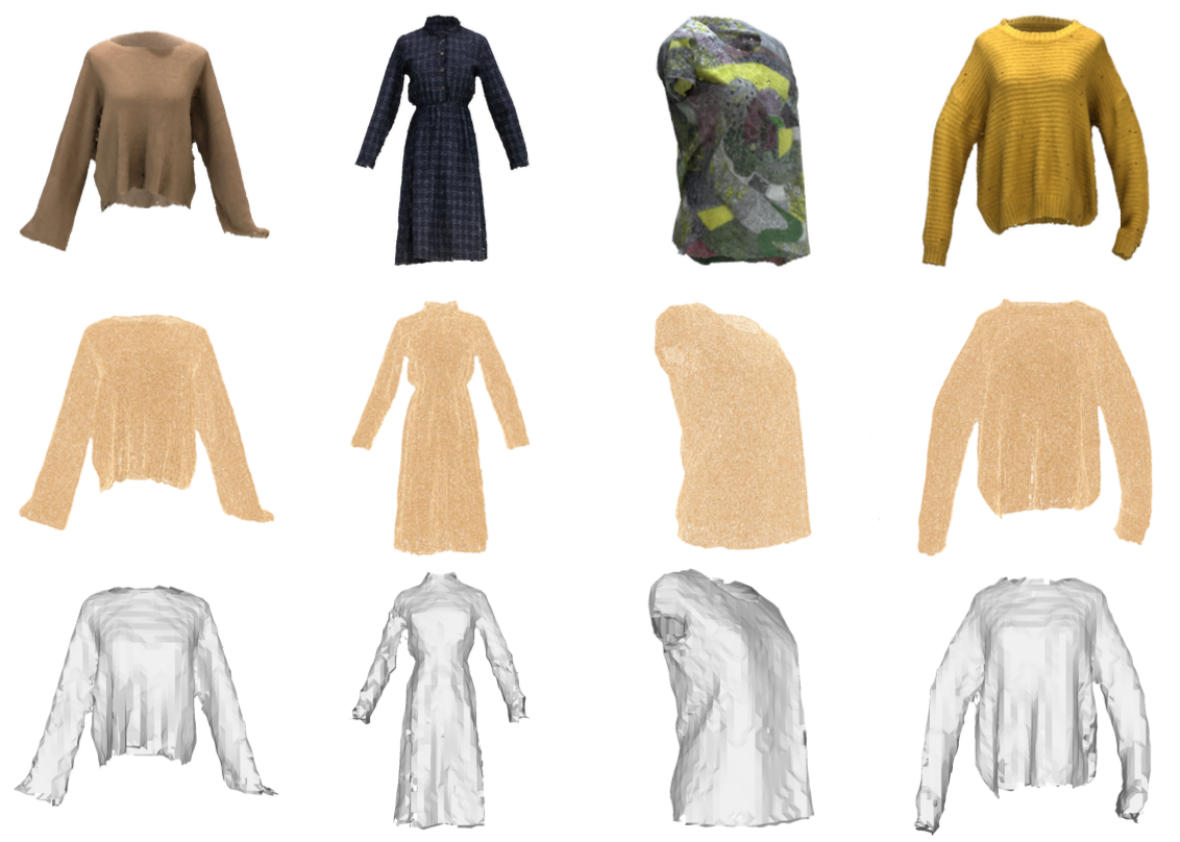}
   \vspace{-20pt}
	\end{center}
		 \caption{\small \textbf{Using our approach to triangulate the outputs of AnchorUDF~\cite{Zhao21a}}, For 4 examples we display the input to the network (a color image), a point cloud of the predicted UDF as originally provided by this network, and a triangulation of the UDF generated using our method.}
	\label{fig:anchorudf}
\end{figure}

%% file: figs/additional_results_rebut.tex

\begin{figure}[t]
	\begin{center}
	 \includegraphics[width=0.65\textwidth]{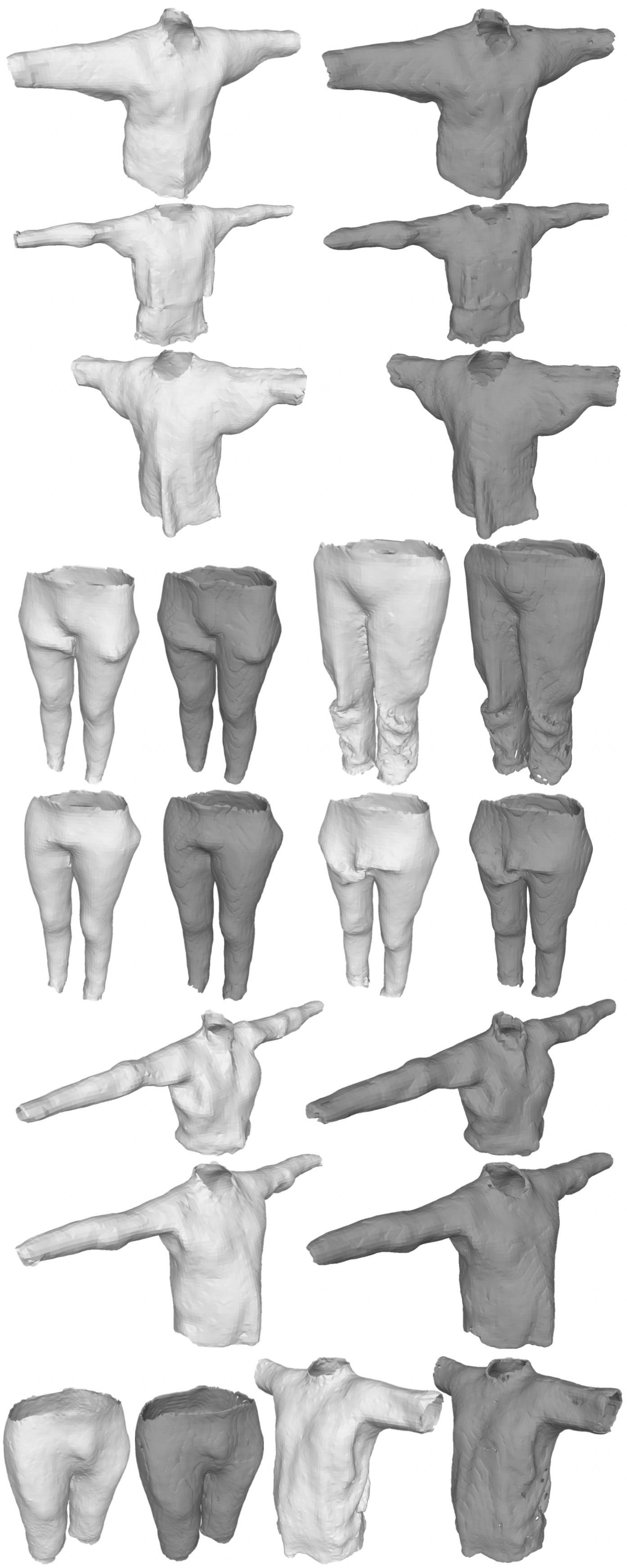} \\
	\end{center}
	\vspace{-5mm}
		\caption{\small \textbf{Comparison of UDF meshing methods:} Comparison of UDF meshing methods: In light grey are our open surface reconstructions ; In dark grey we display the $\epsilon$-inflated baseline that yields wrongly inflated meshes.}
		 \label{fig:additional_results_rebut}
	\vspace{-10pt}
\end{figure}